\documentclass{jov}

\usepackage{booktabs}
\usepackage{dcolumn}
\usepackage{multicol}
\usepackage{subfig}
\usepackage[symbol]{footmisc}
\usepackage{pgfplots}
\usepgfplotslibrary{statistics}
\pgfplotsset{compat=1.8}

\graphicspath{{./images/}}

\begin{document}

\title{CrowdFix: An Eyetracking Dataset of Real Life Crowd Videos}

\abstract{Understanding human visual attention and saliency is an integral part of vision research. In this context, there is an ever-present need for fresh and diverse benchmark datasets, particularly for insight into special use cases like crowded scenes. We contribute to this end by: (1) reviewing the dynamics behind saliency and crowds. (2) using eye tracking to create a dynamic human eye fixation dataset over a new set of crowd videos gathered from the Internet. The videos are annotated into three distinct density levels. (3) Finally, we evaluate state-of-the-art saliency models on our dataset to identify possible improvements for the design and creation of a more robust saliency model.}

 \author{Tahira*}{Memoona}
  {National University of Sciences and}
  {Technology (NUST), Islamabad, Pakistan}
  {http://}{mtahira.mscs17seecs@seecs.edu.pk}

 \author{Mehboob*}{Sobas}
  {National University of Sciences and}
  {Technology (NUST), Islamabad, Pakistan}
  {http://}{smehboob.mscs17seecs@seecs.edu.pk}
 
 \author{Rahman}{Anis U.}
  {National University of Sciences and}
  {Technology (NUST), Islamabad, Pakistan}
  {http://}{anis.rahman@seecs.edu.pk}
 
 \author{Arif}{Omar}
  {National University of Sciences and}
  {Technology (NUST), Islamabad, Pakistan}
  {http://}{omar.arif@seecs.edu.pk}

\renewcommand{\thefootnote}{\fnsymbol{footnote}} 
\footnote[1]{Both authors contributed equally}

\keywords{Eye movement, Eye tracking, Attention, Region of interest, Saliency, Crowd analysis, Fixations}

\maketitle

\section{Introduction}
\label{sec:intro}

Saliency studies form the intersection between natural and computer vision. A quantitative study of saliency provides a structured insight into the human mind on what it perceives to be important in a scene. Visual attention then guides gaze to focus on and further explore that region of interest. To achieve near human accuracy in predicting gaze locations, Saliency models need to be able to approximate gaze over a wide variety of stimuli ~\cite{borji2019saliency}. We approach this problem in two ways: first we discuss static and dynamic stimuli used for modelling saliency as well as the need for specialized datasets to boost saliency modelling. Traditionally most of the active research has been on images, but in the recent years, using dynamic content as the subject of saliency studies has picked up pace. The pace of this research is determined by publicly available, diverse datasets of videos covering multitudes of natural scenes. Datasets such as DIEM ~\cite{mital2011clustering}, HOLLYWOOD-2 ~\cite{mathe2014actions}, and UCFSports ~\cite{mathe2014actions}, LEDOV ~\cite{jiang2018deepvs} and DHFK ~\cite{wang2018revisiting} are dynamic datasets that cover a range of natural scenes.  However, there is an obvious gap for specialized datasets targeting a category of natural scenes. Our study focuses on the category of crowded scenes because it presents an interesting use case: The number of stimuli competing for attention in crowd scenes are larger in number and the crowd activity is far more random and attention grabbing than normal scenes containing one or two object of interest ~\cite{yoo2016visual}. This insight proves useful for monitoring, managing and securing crowds ~\cite{gupta2014design}. To date, there has been only one crowd saliency dataset, namely EyeCrowd, consisting of 500 natural images ~\cite{jiang2014people}. 

Our research contributes by adding a first saliency dataset of crowd videos called 'CrowdFix' and its corresponding saliency information to the pool of publicly available saliency datasets. Crowdfix is a real-life, moving crowds high definition (720p) videos dataset collected in in RGB. Eyetracking results benefit from higher quality datasets ~\cite{vigier2016new}. For this reason we chose not to include videos from pre-existing crowd video datasets due to the lower quality of those videos, i.e. below 720p. The dataset has been further annotated into three different crowd density levels to facilitate understanding of attention modulation within different each level. This also helps in the producing better, more generalized saliency models, particularly deep models by providing a finer categorization of salient images and videos ~\cite{he2019understanding}. We assess the attentional impact of different levels of these crowds on individuals and further evaluate three state-of-art deep learning based saliency models on our datasets to judge how well general saliency models perform for crowd saliency prediction. This analysis serves as a baseline for future design of a crowd saliency model. 

\subsection{Related Work}

Gaze, a synchronized act of the eyes and head, has frequently been used as an intermediary for attention in natural conduct. For example, a human or a robot has to cooperate with contiguous objects and regulate the gaze to accomplish a task while moving in the surroundings. In this sense, gaze control involves vision, response, and attention concurrently to achieve sensory-motor arrangement necessary for the preferred behavior (e.g., reaching and grasping)~\cite{borji2012state}.

Our human visual system is designed to automatically filter the incoming visual information from our gaze. This is done passively based on verdict of visual attention. Visual attention is a mechanism that intervenes between competing aspects of a visual scene and assists in selecting the most important regions while diminishing the importance of others. It is vital to understand how visual attention works to determine how our vision will be directed to the objects presented in front of it. ~\cite{jiang2014people}. Attention is a umbrella term which includes all factors that influence selection. The active selection is expected to be suppressed by two major channels called bottom-up and top-down control. Bottom-up attention is spontaneous attention. It is fast, uncontrolled, and stimulus-driven. Our attention is naturally drawn to salient regions in visual field. The term "salient" is interchangeably used for bottom-up attention ~\cite{borji2012state}. Human visual attention is supposed to look at the salient stimuli in the environment. 

\subsection{Crowds and Visual Attention}

Crowds represent a unique challenge for visual attention selection. A crowd is a big cluster of people assembled together and has attributes like density and movement. Crowds exhibit a distinct category of scenes. Crowd scenes can be categorized as complex scenes, like cross-streets in which several objects interconnect with each other that consists of different movement patterns, for example walking straight and then turning left~\cite{yoo2016visual}. We know that the crowds have an impact on the attention of an individual and can be tested by relating the physical stimuli with the contents of consciousness~\cite{mancas2010dense}. We can then further correlate the different crowd levels with the visual attention to learn the behaviour of an individual while free viewing real life crowds. This information is fundamental for crowd management, safety and surveillance in handling and avoiding emergencies due to rush and congestion~\cite{chiappino2015bio}. Analysis of complex situations like dense crowds can extremely benefit from algorithms which can encode human attention~\cite{mancas2010dense}. This serves many applications in human computer interaction, graphics and user interface design, particularly for small displays, by comprehending where humans look at in a scene. Additionally, knowledge of visual attention is beneficial for automatic image cropping, thumb nailing, image search, image and video compression.~\cite{judd2009learning}.

\subsection{Computational Models for Visual Attention}

Older models integrated complicated characteristics of the Human Visual System (HVS) and and reconstruct the visual input through hierarchically combining low level features. The bottom-up mechanism is the maximum occurring feature found in these models~\cite{le2006coherent}. The core indication which implicates bottom-up attention is the uncommonness and distinction of a feature in a given circumstance~\cite{mancas2010attention}. Bottom-up use a feed-forward method to process visual input. They apply sequential transformations to visual features collected over the entire visual field, to highlight regions which are the most attention-grabbing, significant, eye-catching, or so-called salient information~\cite{borji2012state} However, the existing models of visual attention present a reductionist of visual attention. This is because fixations are not only influenced by bottom-up saliency as determined by the models, but also by various top-down influences. Consequently, comparing bottom-up saliency maps to eye fixations is demanding and requires that one attempts to minimize top-down impacts.~\cite{volokitin2016predicting} One way is to focus on early fixations when top-down influences have not yet come into affect, such as by use of jump cuts in videos, in our case, and MTV style video stimulus~\cite{carmi2006role}.

\section{Our Contribution: The CrowdFix Dataset}
\label{sec:db}

In the only crowd eye tracking experiments that have been done before, images were used. There is no HD (720p), FHD (1080p) or 4K crowd video dataset that exist instead all of the existing datasets have low resolutions. A higher quality dataset leads to better eye fixation information. This is because HD and FHD show a better level of detail in the video and allows for more possibilities of visual exploration.~\cite{vigier2016new}. Most datasets cater exclusively to high-density crowds or abnormal crowds which established the need to have a diversity in the dataset according to crowd density levels. To the best of our knowledge, no such categorization has been performed on existing crowd video datasets.

We collected a crowd dataset consisting of videos that depict real life scenarios. The dataset is categorized into three distinct density levels of the crowds named as sparse, dense free-flowing and dense congested. This dataset is built for studying the influence and saliency in crowds. Therefore, this dataset consists of diverse real-life, moving crowds. It has a total of 89 videos cut into 434 clips for MTV style videos. Having high resolution of the videos as the starting key point, our dataset comes with the resolution of 1280$\times$720 with 30 frames per second. None of the videos in this dataset are taken from any previously existing datasets. For maintaining the clarity and simplicity of the videos, none of them is in a fast forward motion nor any of them has a watermark on it while all of the videos being in RGB. For generating the dataset we picked the crowd videos under the Creative Commons depicting multiple real life crowded scenes. The stimulated crowd videos were not considered at all. We collected a wide variety of moving crowd scenes while assessing the varying densities of crowds. The videos were then later finalized. The categorization of the crowd density levels is concluded from the results of the participants. The major step in creating the stimulus was to maximize the bottom up attention. Since bottom up attention is the involuntary attention it follows that the stimulus should change frequently and abruptly. In terms of videos this can be achieved by using jump cuts by combining videos of a very short length back to back together. We call these very short videos as a 'clip'. Based on the research of ~\cite{carmi2006role}, each snippet duration varies from 1 second to up to 3 seconds. Any clip of length greater than this would invoke top down attention. To create the clips from the crowd videos we take all the videos from each density level and randomly shuffle them. This ensures there is no sequence based on the crowd density to make it predictable. Each video has a duration of 1 - 3 seconds. These snippets are again randomly combined into two videos of approximately 10 minutes each. We then present the stimulus to the participants. Table \ref{tab:datasetsummary} shows the attributes of the real life crowd dataset. 

\begin{table}[!ht]
\footnotesize
\centering
\begin{tabular*}{350pt}{@{\extracolsep\fill}llcccccccccclD{.}{.}{3}l@{\extracolsep\fill}}
\toprule
Attributes      & Details \\ 
\midrule
Stimuli type    & Outdoor daytime/nighttime human moving crowds \\ 
Sources         & 05 (Flickr, Pexels, Pixabay, Vimeo and Youtube)\\ 
Licence         & under Creative Commons \\
Number of videos& 89 video clips \\
Categories      & Sparse, dense free-flowing, and dense congested \\
Videos per category & Sparse (15), dense free-flowing (41), and dense congested (33) \\
Total video frames  & 37,493 \\
Video frame size    & 1280 $\times$ 720 \\
Audio   & No \\
Video snippets  & 485 (1--3s each) \\ 
Video clippet   & Randomly selected snippets \\
Video clippet duration & $\sim$10 mins \\
\bottomrule
\end{tabular*}
\caption{Dataset summary table}
\label{tab:datasetsummary}
\end{table}

\subsection{Dataset Annotation}

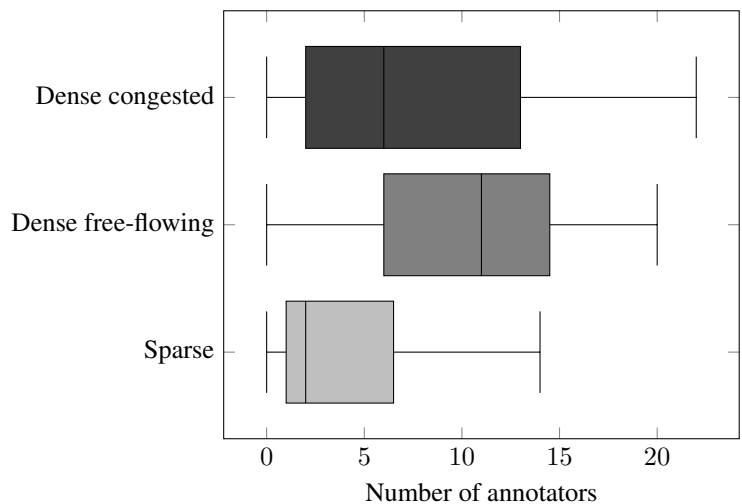
\begin{figure}[h!]
\centering

\begin{tikzpicture}
  \begin{axis}
    [
    ytick={1,2,3},
    yticklabels={Sparse, Dense free-flowing, Dense congested},
    xlabel={Number of annotators},
    ]
\addplot+[boxplot,mark=none,draw=black,fill=black!25]
table[row sep=\\,y index=0] {
data\\
1\\ 12\\ 8\\ 1\\ 5\\ 7\\ 1\\ 4\\ 0\\ 7\\ 1\\ 1\\ 11\\ 0\\ 17\\ 3\\ 4\\ 7\\ 0\\ 1\\ 4\\ 0\\ 1\\ 0\\ 0\\ 0\\ 3\\ 0\\ 16\\ 3\\ 17\\ 2\\ 1\\ 21\\ 6\\ 1\\ 2\\ 5\\ 5\\ 0\\ 6\\ 6\\ 1\\ 0\\ 14\\ 0\\ 1\\ 4\\ 20\\ 21\\ 2\\ 6\\ 2\\ 2\\ 18\\ 16\\ 7\\ 2\\ 0\\ 5\\ 10\\ 1\\ 15\\ 1\\ 0\\ 1\\ 22\\ 1\\ 1\\ 1\\ 1\\ 0\\ 5\\ 13\\ 1\\ 18\\ 0\\ 5\\ 1\\ 0\\ 9\\ 6\\ 3\\ 4\\ 0\\ 6\\ 10\\ 1\\ 7\\ 0\\ 0\\ 1\\ 6\\ 7\\ 0\\ 1\\ 4\\ 10\\ 0\\ 1\\ 3\\ 4\\
};
\addplot+[boxplot,mark=none,draw=black,fill=black!50]
table[row sep=\\,y index=0] {
data\\
20\\ 10\\ 9\\ 15\\ 15\\ 12\\ 8\\ 10\\ 6\\ 14\\ 7\\ 2\\ 11\\ 9\\ 5\\ 16\\ 16\\ 11\\ 8\\ 6\\ 11\\ 15\\ 2\\ 13\\ 4\\ 1\\ 13\\ 15\\ 5\\ 7\\ 4\\ 8\\ 16\\ 1\\ 11\\ 3\\ 19\\ 10\\ 12\\ 13\\ 5\\ 12\\ 17\\ 8\\ 9\\ 3\\ 12\\ 17\\ 1\\ 1\\ 17\\ 16\\ 16\\ 15\\ 5\\ 5\\ 14\\ 13\\ 9\\ 17\\ 11\\ 2\\ 7\\ 6\\ 3\\ 11\\ 0\\ 11\\ 9\\ 7\\ 11\\ 6\\ 15\\ 6\\ 13\\ 3\\ 4\\ 16\\ 13\\ 11\\ 13\\ 9\\ 15\\ 15\\ 6\\ 16\\ 8\\ 5\\ 12\\ 18\\ 5\\ 19\\ 14\\ 14\\ 13\\ 16\\ 16\\ 11\\ 8\\ 10\\ 11\\ 17\\
};
\addplot+[boxplot,mark=none,draw=black,fill=black!75]
table[row sep=\\,y index=0] {
data\\
2\\ 1\\ 6\\ 7\\ 3\\ 4\\ 14\\ 9\\ 17\\ 2\\ 15\\ 20\\ 1\\ 14\\ 1\\ 4\\ 3\\ 5\\ 15\\ 16\\ 8\\ 8\\ 20\\ 10\\ 19\\ 22\\ 7\\ 8\\ 2\\ 13\\ 2\\ 13\\ 6\\ 1\\ 6\\ 19\\ 2\\ 8\\ 6\\ 10\\ 12\\ 5\\ 5\\ 15\\ 0\\ 20\\ 10\\ 2\\ 2\\ 1\\ 4\\ 1\\ 5\\ 6\\ 0\\ 2\\ 2\\ 8\\ 14\\ 1\\ 2\\ 20\\ 1\\ 16\\ 20\\ 11\\ 1\\ 11\\ 13\\ 15\\ 11\\ 17\\ 3\\ 4\\ 9\\ 2\\ 19\\ 2\\ 9\\ 12\\ 1\\ 8\\ 5\\ 4\\ 17\\ 1\\ 5\\ 17\\ 4\\ 5\\ 18\\ 3\\ 3\\ 2\\ 10\\ 6\\ 3\\ 2\\ 15\\ 12\\ 9\\ 2\\
};
  \end{axis}
\end{tikzpicture}
\caption{Categorization distribution by annotators.}
\label{fig:categorydistribution}
\end{figure}

The objective behind dataset annotation is to divide the dataset into distinct crowd density levels. All the previously available crowd videos datasets lacked the density feature. The major attributes of crowds include density, orientation, time and location of event, type of event, demographics and organization within the crowd. Hence we choose to focus on different crowd density levels as well to perform better from social, psychological and computational point of view.  1 - 1.5 humans per square meter is treated as sparse, 2 humans per square meter is treated as dense free-flowing and if 3 - 4 humans per square meter then it is treated as dense congested; this is all done for moving crowds~\cite{crowden}. 23 annotators (5 males and 18 females, in between the age group of 17 and 40) free-viewed the crowd videos. After viewing each video they were given some time to mark the video as one of the level explained in the beginning. Figure~\ref{fig:categorydistribution} shows the distribution of the categories chosen by the annotators that were further assigned to all the videos in the dataset.  

\begin{figure}[!ht]
\centering
\subfloat{
\includegraphics[width=0.30\linewidth]{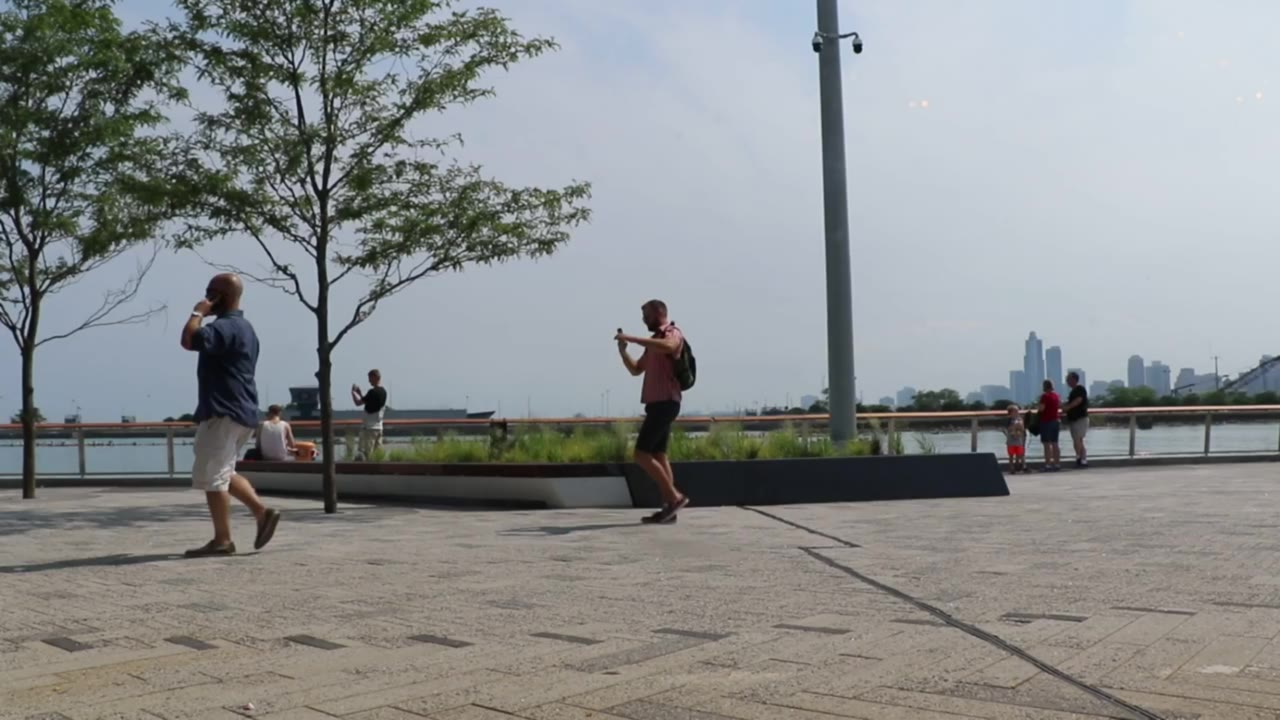}}\quad
\subfloat{
\includegraphics[width=0.30\linewidth]{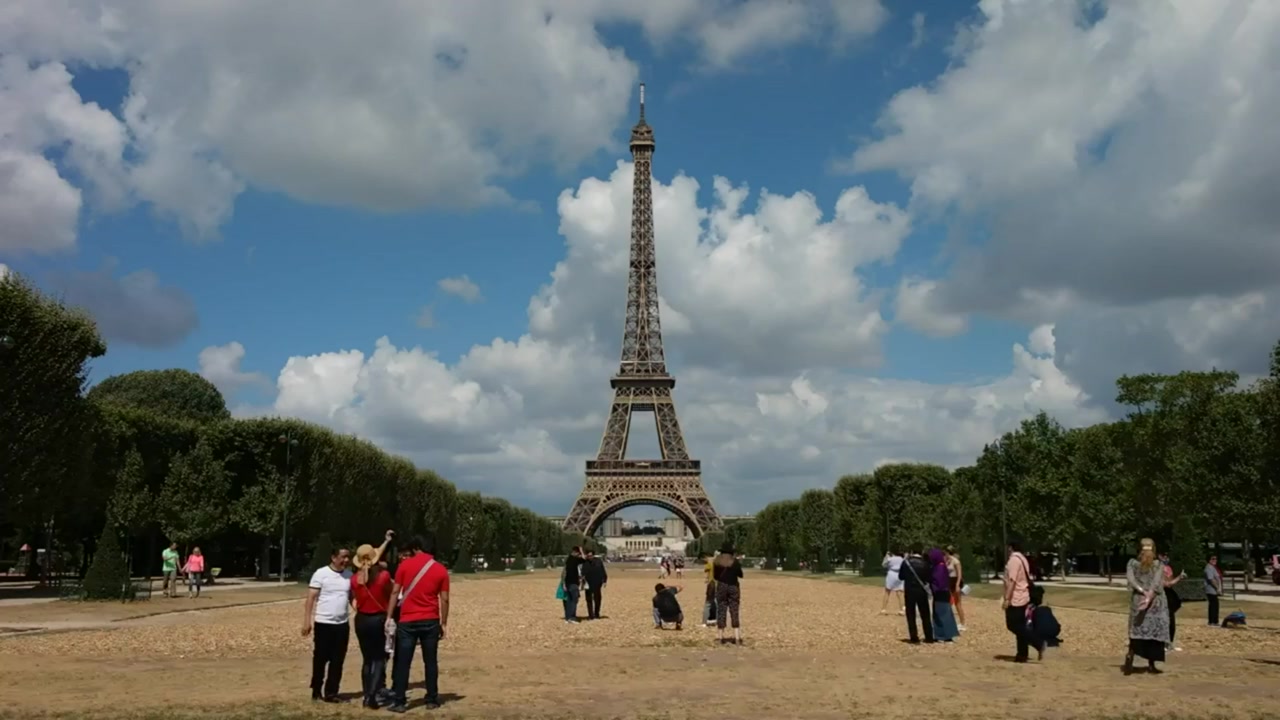}}\quad
\subfloat{
\includegraphics[width=0.30\linewidth]{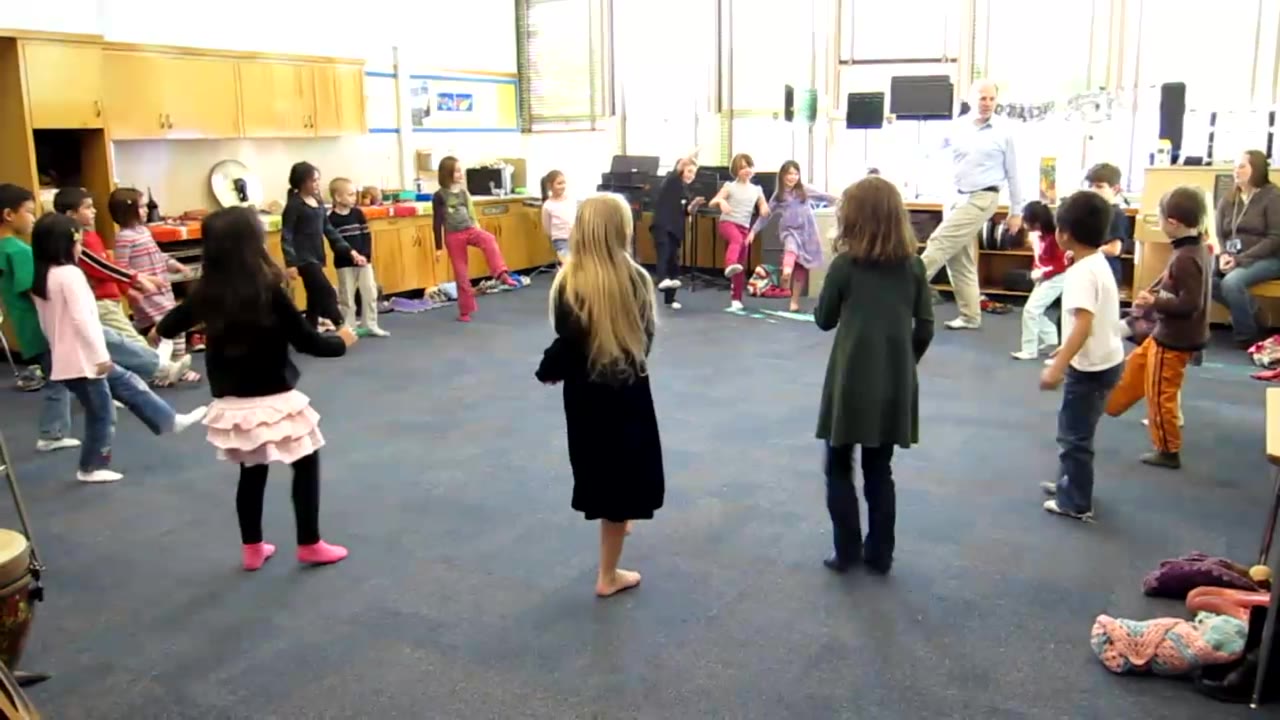}}\\
\subfloat{
\includegraphics[width=0.30\linewidth]{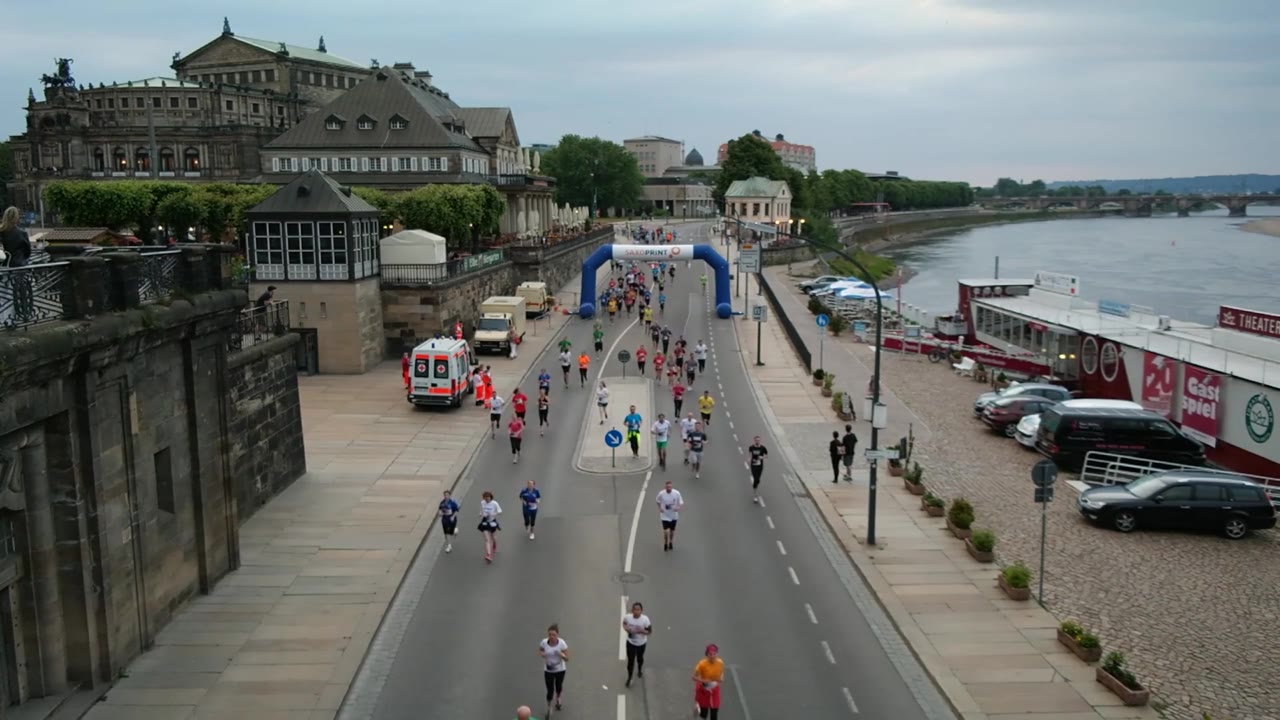}}\quad
\subfloat{
\includegraphics[width=0.30\linewidth]{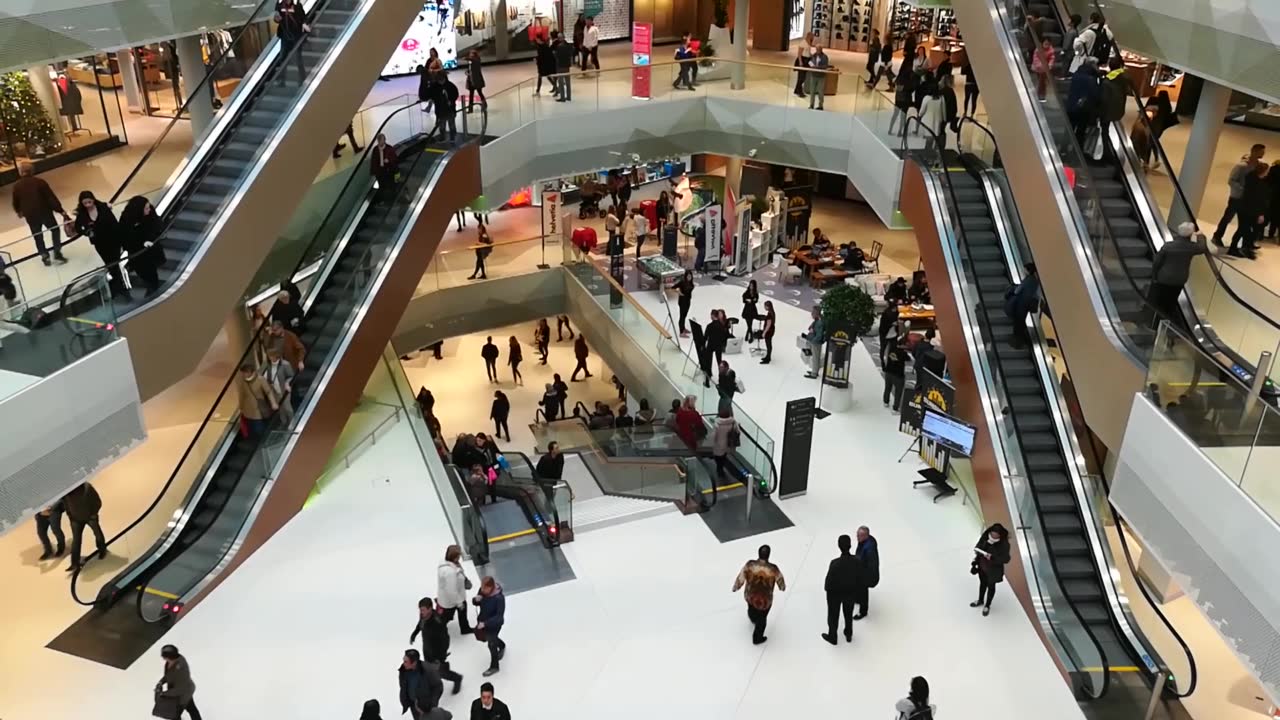}}\quad
\subfloat{
\includegraphics[width=0.30\linewidth]{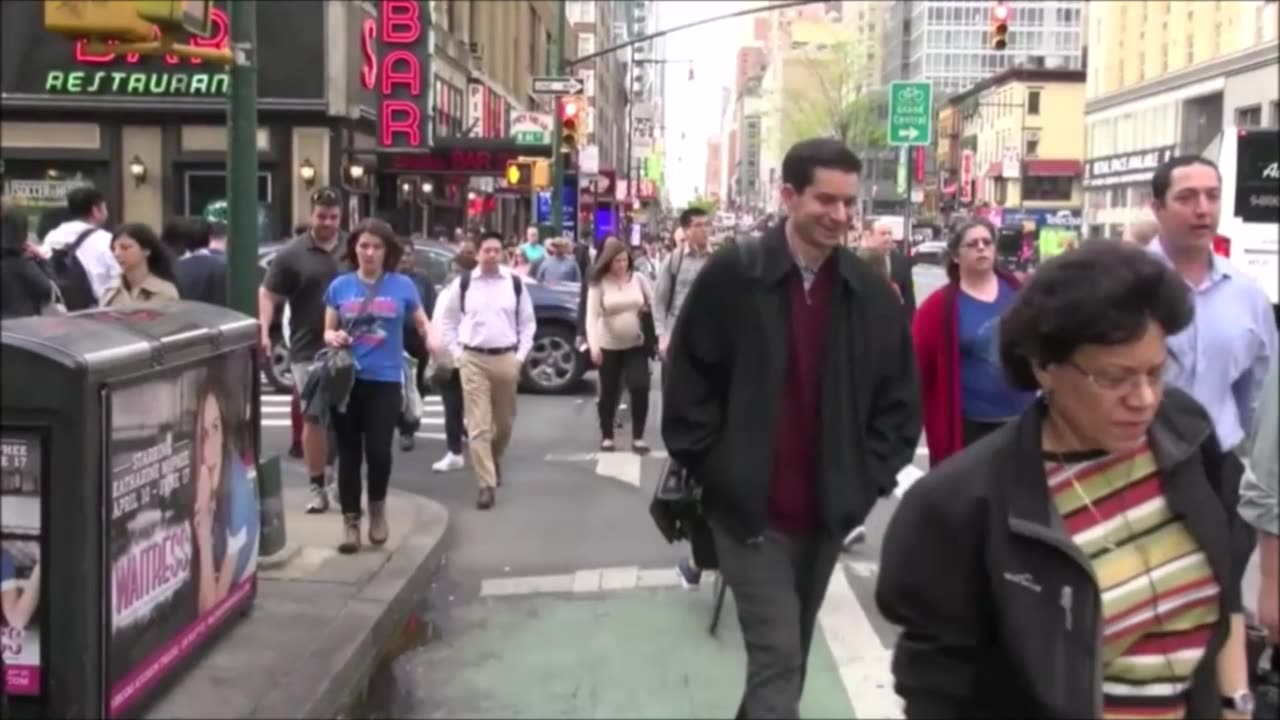}}\\
\subfloat{
\includegraphics[width=0.30\linewidth]{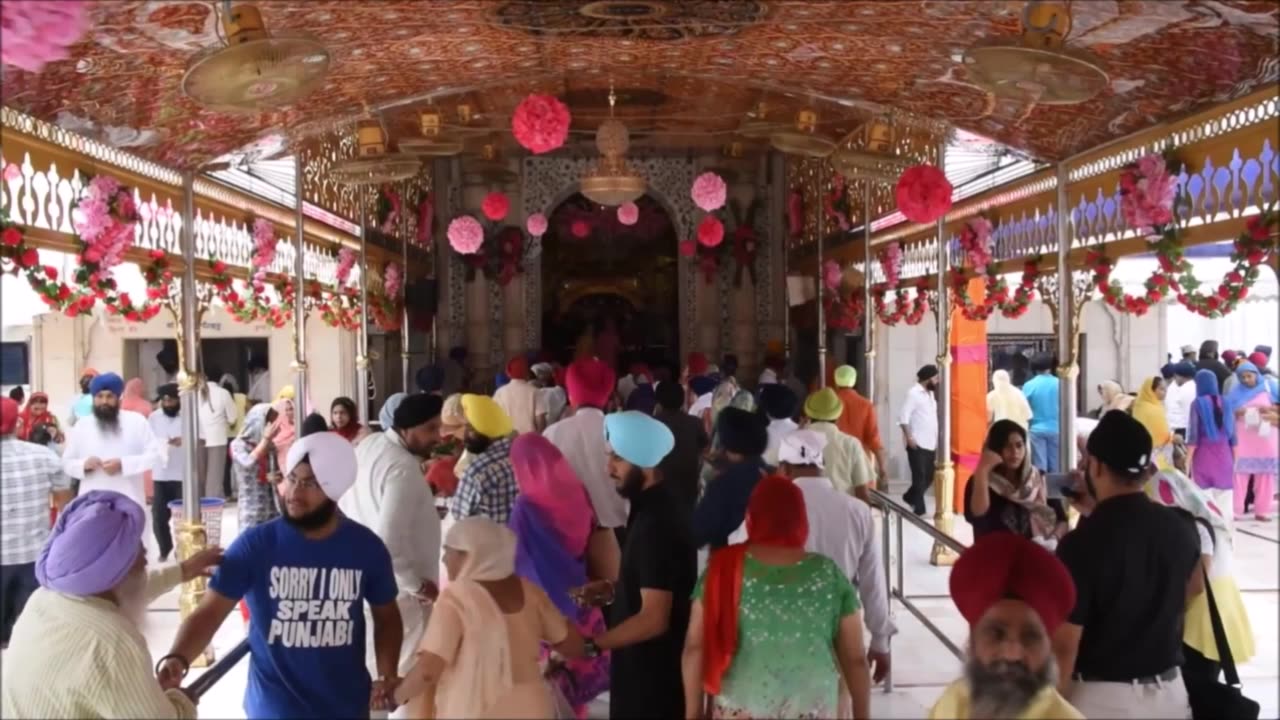}}\quad
\subfloat{
\includegraphics[width=0.30\linewidth]{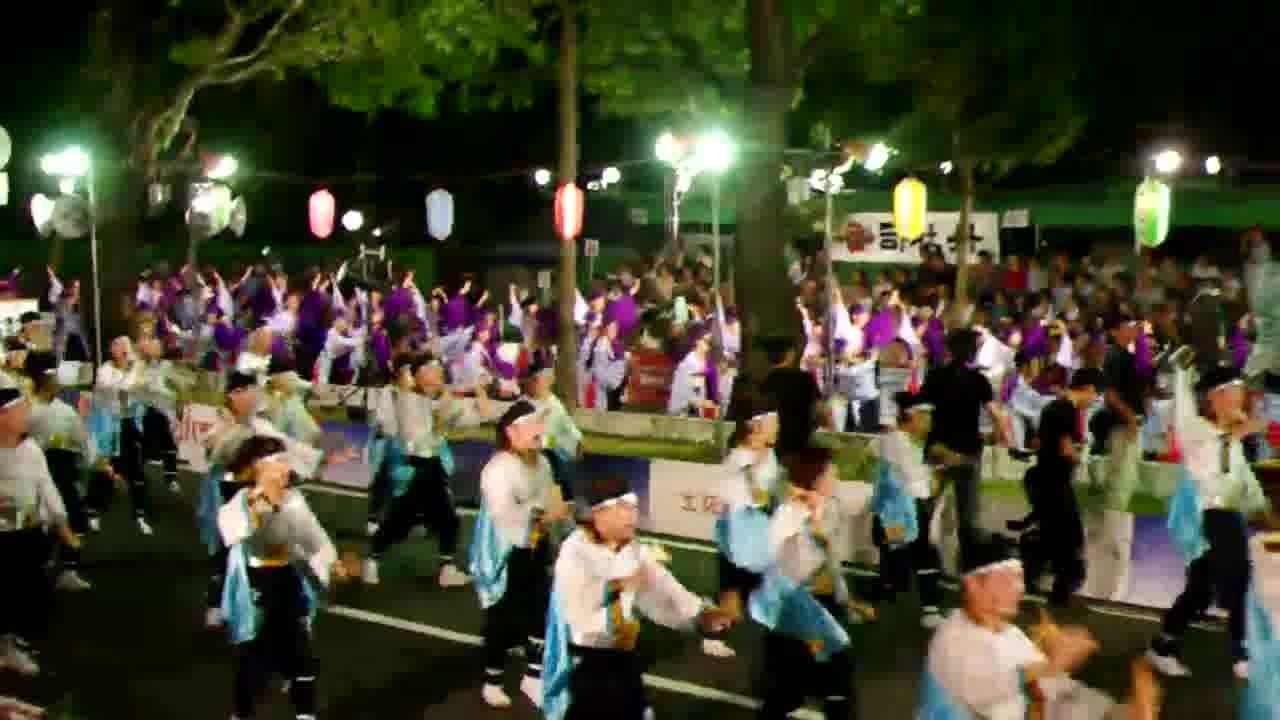}}\quad
\subfloat{
\includegraphics[width=0.30\linewidth]{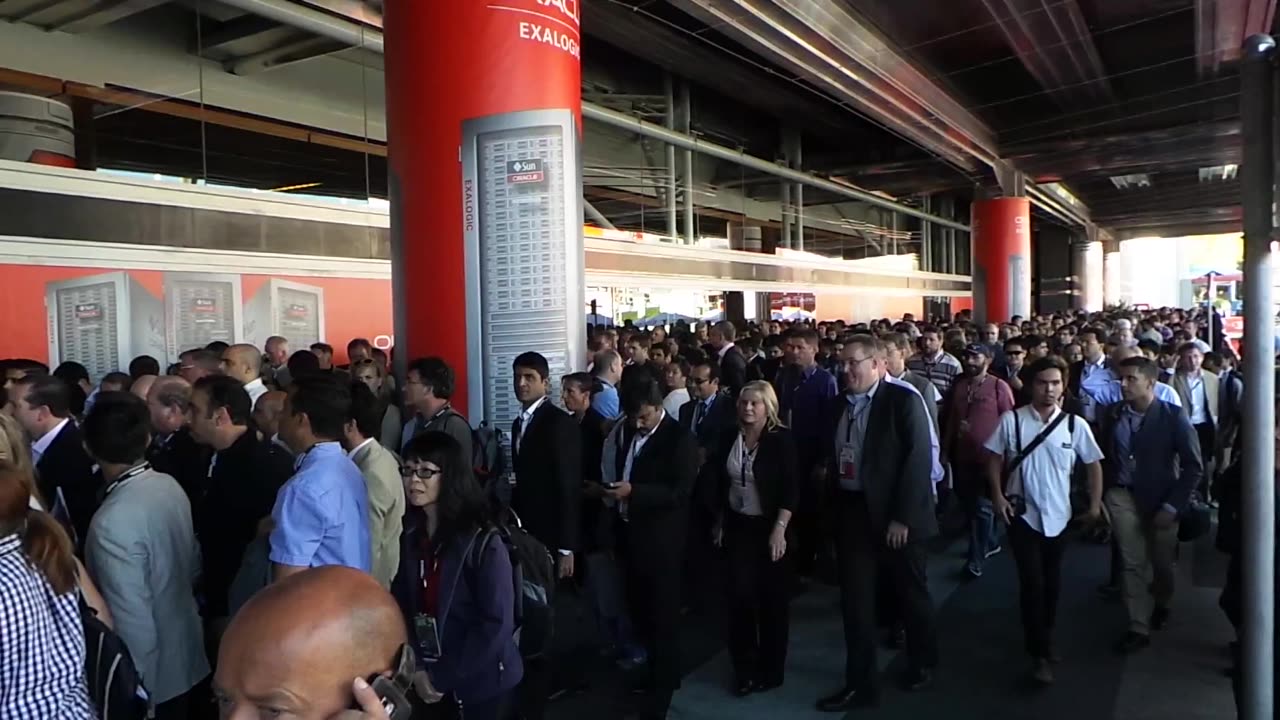}}
\caption{Sample frames for varying crowd density levels. Top row: Sparse. Middle row: Dense and free flowing. Bottom row: Dense and Congested}
\label{fig:sampleframes}
\end{figure}

Since we saved the decision about the density of the video right after showing the video to the participants we can be fairly certain that the participant's judgment was not influenced by other videos. And the participant could pause as long as they wished before moving on to the next video. Figure~\ref{fig:sampleframes} shows the sample images of different levels of crowd density. The rows represent sparse, dense free-flowing, and dense congested crowds respectively.

\subsection{Eyetracking Data Acquisition}

\subsection{Eye tracking, general motivation and process}

We chose eyetracking as our ground truth collection approach to harvest good data. ~\cite{tavakoli2017saliency} Ground truth refers to human eye movement data obtained from real life observers who viewed the stimulus. This also works out well because our stimulus are videos where each frame moves rapidly and only stays on screen for a split second. 

\begin{figure}[h!]
\centering
\includegraphics[scale=0.6]{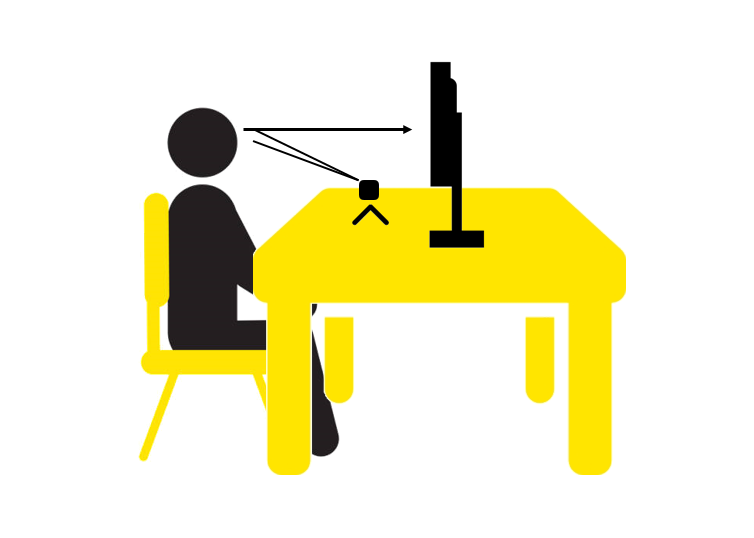}
\caption{Conducting the Experiment with a participant}
\label{fig:experiment}
\end{figure}

Videos were displayed on a 23.8” HP 24es LCD monitor (resolution 1920 x 1080) with the person resting his face on the head and chin rest to minimize any kind of ambiguity and shakiness in eye movement tracking. The distance of the viewer from the screen was kept as 60 cm. 32 participants volunteered for the free viewing of the videos while their gaze points were being recorded. Figure~\ref{fig:experiment} shows the experimental setup and the experiment being performed by a volunteer.

All participants were shown the same set of videos in the same order. Free viewing allows participants to involve in natural visual expedition, while reassuring them to pay solid attention on the screen during the session. Therefore, some instructions kept the participants naive to the objective of experiment. Also, no one had seen the stimulus before. The instructions given were as follows:
\begin{enumerate}
    \item You have to thoroughly watch the videos that are going to be played in front of you
    \item Try to follow the main things in the video as some general questions can be asked at the end of the session
    \item Make sure your eye sight is normal or corrected and you’re not wearing any polarized glasses or mascara
\end{enumerate}

The EyeTribe eye tracker is used to perform the experiment with our dataset. The company 'The Eye Tribe' endorses their eye tracker to be "the world's first 99 eye tracker with full SDK". It has two software suites that supplements the device i.e. EyeTribe UI and EyeTribe Server.  It has a sampling rate of 60 Hz and standard precision of 0.5$^\circ$ to 1.0$^\circ$. Eye tracking is a measurement of eye movement or activity. Near infra-red light is fixed towards the focal point of the eyes namely pupils, instigating visible reflections in the cornea (the outer-most optical part of the eye), and tracked by a camera. The results provides us with the fixation data that is a time in which our eyes are locked towards a particular object in a visual angle~\cite{dalrymple2018examination}. 

An eye tracker's efficiency is commonly assessed by two metrics: accuracy and precision. Systematic error or accuracy replicates the eye tracker's capability to assess the point of regard. It is also defined as the mean difference between a test stimulus position and the measured gaze position~\cite{holmqvist2012eye}. Whereas the precision invokes the eye tracker's ability to deliver steady measurements, and is appraised by calculating the root mean square noise~\cite{holmqvist2012eye}.

The eye tracker is controlled by the python based PyGaze toolbox (an alternative of the Psychtoolbox from the MATLAB) script running on Lenovo 320-15IKB (Intel Core i7-8550U CPU- @ 1.80 GHz, 8 GB, Windows 10), using a HP 24es LED monitor (23.8 inch, 60 Hz, 1920 $\times$ 1080 pixels, with dimensions 52.7x29.6 cm degrees of visual angle). Calibrations are performed using a nine point grid scripted in Python. Table~\ref{tab:properties} shows the the properties of the eye tribe eye tracker used for the experiment.

\begin{table}[!ht]
\footnotesize
\centering
\begin{tabular*}{350pt}{@{\extracolsep\fill}llcccccccccclD{.}{.}{3}l@{\extracolsep\fill}}
\toprule
Property                & Value \\ 
\midrule
Eye tracking principle & Non-invasive, image based eye tracking \\ 
Sampling rate          & 30 Hz or 60 Hz \\
Accuracy               & 0.5--1.0$^\circ$ \\
Spatial resolution     & 1.0$^\circ$ (RMS) \\
Latency                & $<$20 ms at 60 Hz \\
Calibration            & 9, 12 or 16 points \\
Operating range        & 45--75 cm \\
Tracking area          & 40$\times$30 cm at 65 cm distance \\
Gaze tracking range    & Up to 24" \\
API/SDK                & C++/C\#/Java \\
Data output            & Binocular gaze data \\
\bottomrule
\end{tabular*}
\caption{Eye tracker Properties}
\label{tab:properties}
\end{table}

To establish the aforementioned measures, gaze position is recorded during two-second periods of fixating a target stimulus. The targets appear consecutively, with an inter-trial interval of one second, on locations that were different from the calibration grid. The target grid spans 25.81 degrees of visual angle horizontally, and 19.50 degrees vertically (centred around the display centre). This is done to ensure that the tracker is feasible enough for performing the experiment in terms of: systematic error (spatial accuracy in degrees of visual angle), precision (Temporal accuracy in degrees of visual angle), and sampling accuracy.

\subsection{Experimental design}
We use convenience sampling for conducting the eye tracking experiment. This means that we search for volunteers amongst colleagues and people around us in the university only. Data cleaning is also performed by comparing the results from calibration and validation from the accuracy and precision that were calculated at both the times. It regards the systematic error of less than 1.7$^\circ$ as being acceptable~\cite{blignaut2014eye}. Based on the data cleaning process 6 participant's data was discarded, leaving us with 26 participants - 16 females and 10 males aged between 17 - 40 years. Since the research was being held at the graduate level hence the participants were mostly graduates with normal or corrected vision. Since eye tracking falls under human behavioural research, we choose elements from commonly used behavioural experiments. The design of the experiment is motivated by the need to quantify the response of participants to the stimuli in an objective and reliable way. To maintain reliability, the stimulus duration and sequence for each participant is fixed. The experiment is divided into two identical blocks with a break of 3 - 5 minutes and starting again with a re-calibration process. The video sequence within each block remains the same for each participant. The MTV style sequence in each of the blocks keep the stimulus unpredictable and preserves the objectivity. On the same note, the stimuli did not overlap as it is a mixed design. It includes longitudinal data (by collecting a sample at the rate of 60 Hz) and cross-sectional data across several participants. The hypothesis of underlying the design of the eye tracking experiment is defined by cause, effect and goal. The cause is our stimuli, that is the crowd videos shown on a monitor. The effect is the change in the visual attention of the participant. And out goal is to analyze visual attention in crowd videos. The independent variable in our experiment are the crowd videos. The well defined density levels of a crowd ensures that we provide sufficiently diverse stimuli to our participants. The dependent variables therefore, are the raw gaze data and fixation data. To allow the fixation data to accurately represent actual eye fixations that rest on salient regions only however, is tough and requires the attempt to reduce top-down influences by concentrating on initial fixations on a stimulus~\cite{volokitin2016predicting}. One way to keep the focus on early fixation is the use of jump cuts in videos, in our case, and MTV style video stimulus. This is in line with the understanding that salient parts of a scene consist of an unexpected commencements or local singularity~\cite{le2006coherent}. Early attention is learnt from initial interactions, later viewing involves task/memory and other complex processes. Hence the reassembling is done into two MTV style videos named MTV1 and MTV2 of 10:12 and 10:37 minutes of duration respectively. These videos for bottom up attention helps in reducing the time for a participant to think. Therefore, the recorded data is objective.

\subsection{Database Location, Structure and Accessibility}

The dataset is organized into stimuli containing the video frame, the fixation maps which is a binary map pinpointing the exact location of the fixations, and saliency maps that are the Gaussian blurred fixations. The sigma for the Gaussian was set equal to one degree of visual angle according to standard procedure, which was 38 in our dataset.

A brief description of the dataset along with the download link can be found at: https://github.com/MemoonaTahira/CrowdFix.

\section{Results and Analysis}

\subsection {Fixation Overview}

All the resulted log files from the eye tracking experiment had the data of participant's fixations across all the videos. We performed general analysis on those log files to come up with some information about the experiment. Total number of fixations and average fixations were computed from MTV1 and MTV2 both. The count of minimum and maximum number of fixations was also calculated on both the parts. The values clearly show that MTV1 has more number of fixations than MTV2 therefore having more average fixations in the first part as well. Table~\ref{tab:attributes} shows the results of fixation data gathered by all the participants in MTV part 1 and 2.

\begin{table}[!ht]
\footnotesize
\centering
\begin{tabular*}{450pt}{@{\extracolsep\fill}llcccccccccclD{.}{.}{3}l@{\extracolsep\fill}}
\toprule
Attributes              & MTV 1         & MTV 2  \\ 
\midrule
Total fixations         & 11829         & 11518 \\
Average fixations       & 454.9615      & 443 \\
Median of fixations     & 455     	    & 443 \\
Maximum fixations       & 455  	        & 443 \\
Minimum fixations       & 454  	        & 443 \\
\bottomrule     
\end{tabular*}
\caption{Fixations Data on MTV1 and MTV2}
\label{tab:attributes}
\end{table}

\subsection{Gaze Data Visualization}

\begin{figure}[!ht]
\centering
\footnotesize
\begin{tikzpicture}
  \begin{axis}[
    xbar, xmin=0, xmax=180,
    bar width=15pt,
    width=0.4\linewidth, 
    height=6cm, 
    enlarge x limits=0.1,
    enlarge y limits=0.8,
    xlabel={Number of videos},
    symbolic y coords={Sparse, Dense free-flowing, Dense congested},
    ytick=data,
    nodes near coords, nodes near coords align={horizontal},
    ]
    \addplot [draw=black,fill=black!25] coordinates {(116,Sparse) (163,Dense free-flowing) (155,Dense congested)};
  \end{axis}
\end{tikzpicture}
\caption{Crowd density levels against number of videos}
\label{fig:categoryvsvideos}
\end{figure}
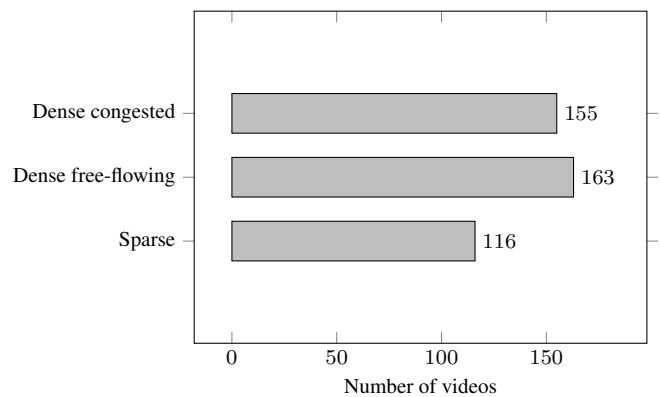

The final analysis is done with respect to crowd density levels against their number of videos.Figure~\ref{fig:categoryvsvideos} shows the distribution of videos over different levels of crowds. Each level has different number of videos as presented in the figure.

Different density levels of the crowd were evaluated on two things being, number of fixations and duration of fixation. Table~\ref{tab:insights} shows the results of the evaluation on sparse, Dense free-flowing and dense congested crowd levels. From the levels that we have, dense congested has the highest number of fixations since there are more people to look at as compared to dense free-flowing and sparse. But even if sparse has less number of fixations, it has the highest duration of fixations on the screen.  

\begin{table}[!ht]
\footnotesize
\centering
\begin{tabular*}{450pt}{@{\extracolsep\fill}llcccccccccclD{.}{.}{4}l@{\extracolsep\fill}}
\toprule
Crowd Level                 & Number of Fixations   & Average Fixation Duration  \\ 
\midrule
Sparse                      & 3016	              	& 232.3886 ms  \\
Dense free-flowing          & 4238	                & 229.4085 ms  \\
Dense congested             & 4030	             	& 228.7178 ms  \\
\bottomrule     
\end{tabular*}
\caption{Analysis Over Crowd Density Levels}
\label{tab:insights}
\end{table}

Fixation duration explains for how long the fixation of an individual lied on the screen. Hence, sparse having the highest fixation duration shows us that it catches most attention of the individuals looking at the crowd as they have lesser people to look at so they spend longer time on viewing such scenes as compared to dense free-flowing and dense congested.

\begin{figure}[!ht]
\centering
\subfloat{
\includegraphics[width=0.30\linewidth]{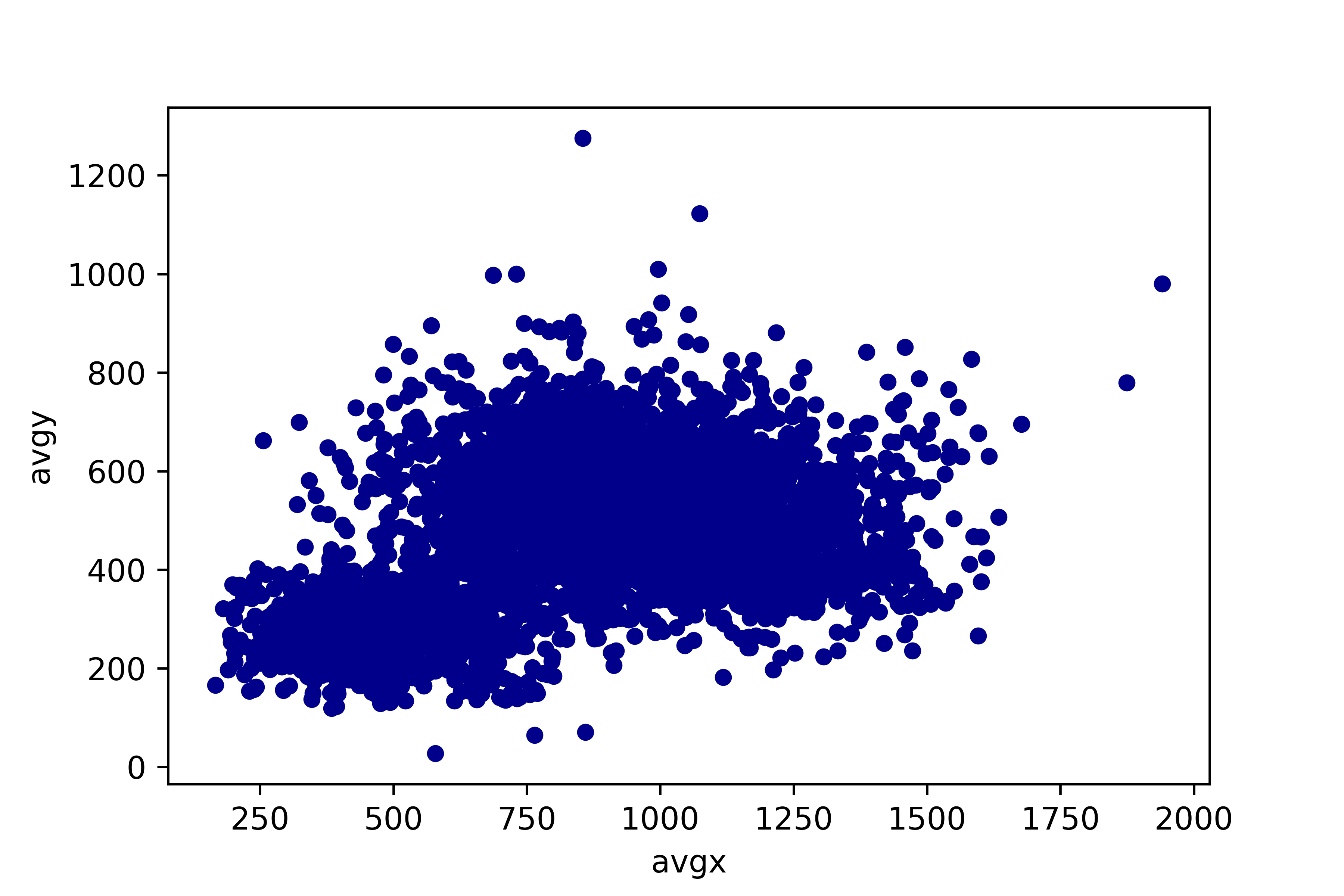}}\quad
\subfloat{
\includegraphics[width=0.30\linewidth]{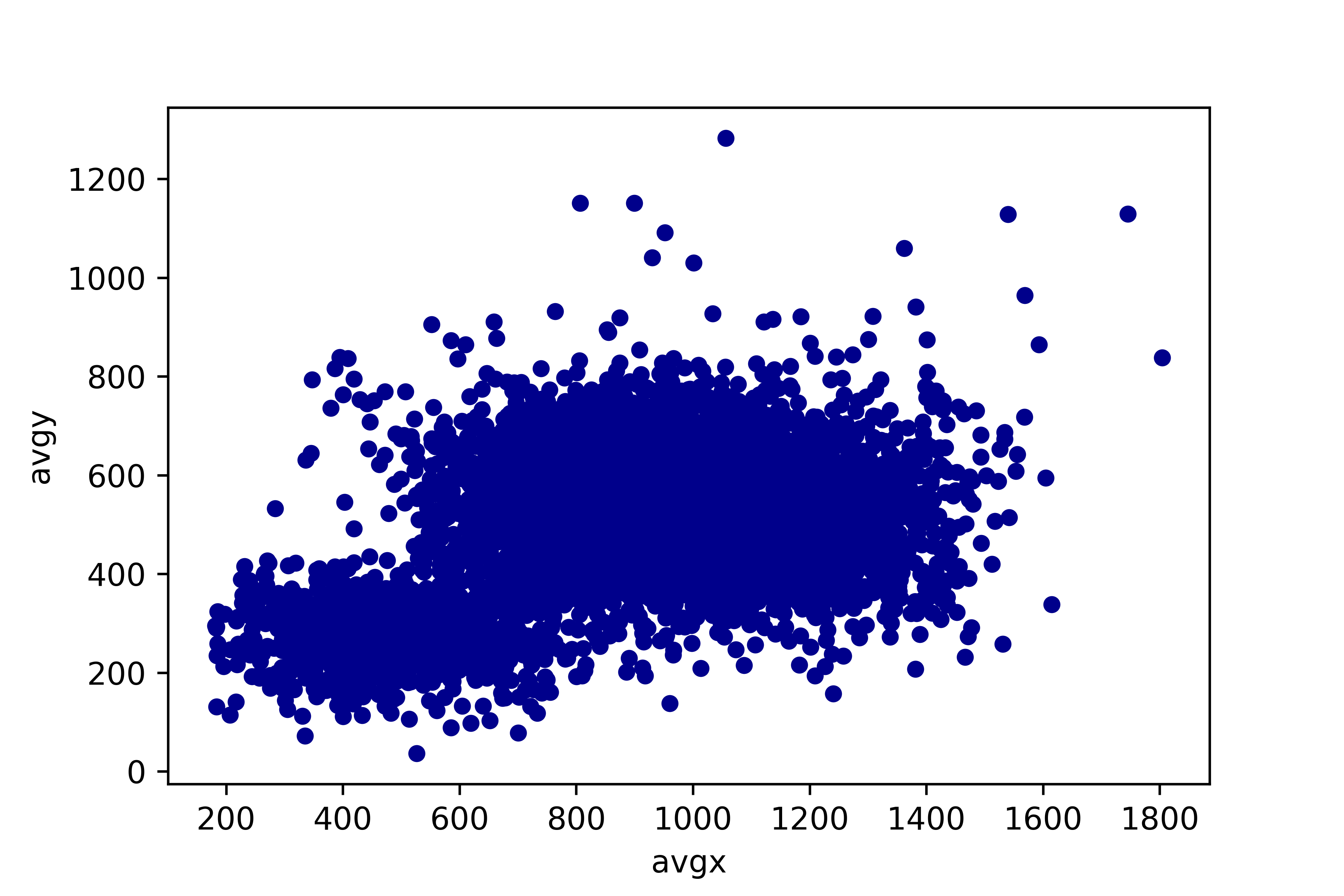}}\quad
\subfloat{
\includegraphics[width=0.30\linewidth]{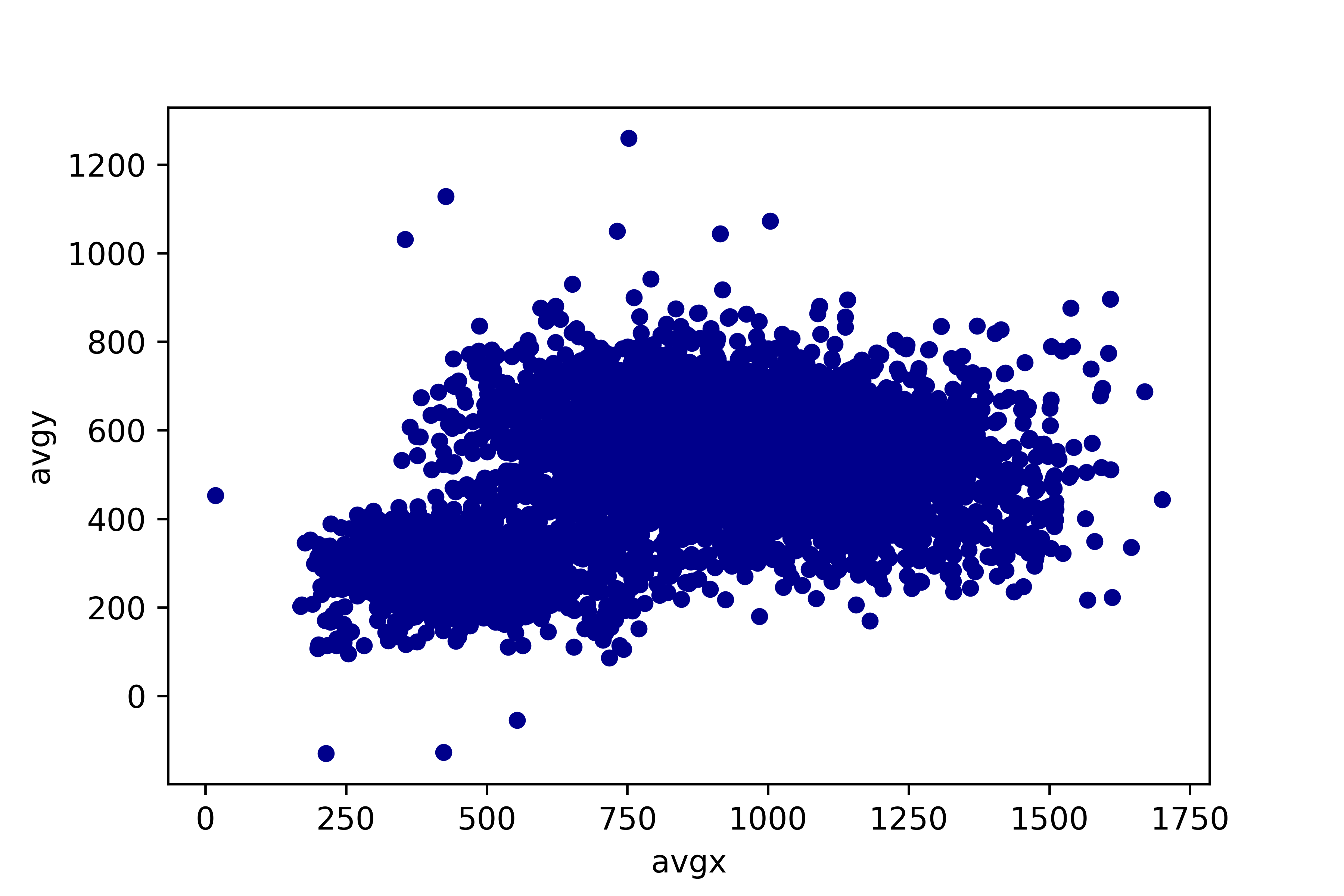}}\\
\caption{Fixation Locations of participants in Different Crowd Density Levels}
\label{fig:fixations}
\end{figure}

Fixation location is also one of an important aspect to look at while interpreting the results. It reveals the areas where the participants fixate on the screen. The images below are the graphs representing fixation locations across all the participants of different crowd levels. It can be seen that all the fixations lie closer to the center and form a big cluster with all the points tightly loaded. Figure~\ref{fig:fixations} shows the fixation locations of the participants throughout the experiment on different crowd density levels. It represents the graphs for sparse, dense free-flowing and dense-congested categories from left to right respectively.    

\begin{figure}[h!]
\centering
\includegraphics[width=0.4\linewidth]{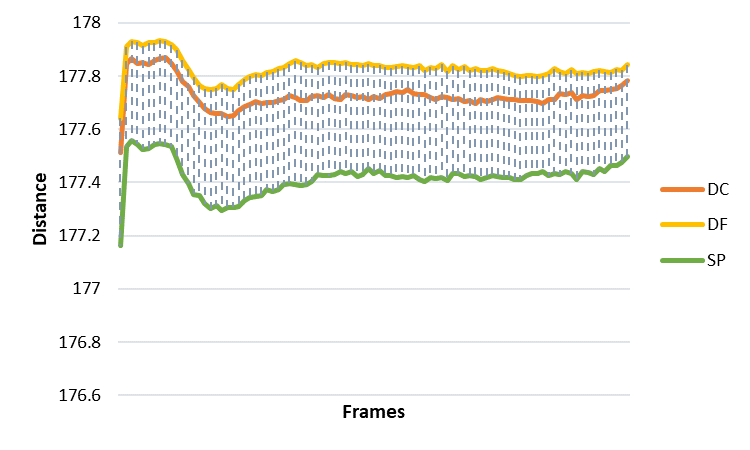}
\caption{Distance of Fixations from the Center}
\label{fig:alldist}
\end{figure}

The fixation coordinates of all the participants were extracted and used for calculating the distance from the center. Figure~\ref{fig:alldist} shows the results of different crowd levels. A peak at the first frame can be seen due to the bottom-up saliency influences. The fixations from sparse density level seem to be more close to the center as there are lesser things to be looked at in those scenes that revolve around the center of the screen. Hence the lesser entities in the scene catch attention for longer periods of time being consistent towards the center of the screen. While the fixations from dense free-flowing and dense congested seem to be more distributed than sparse. Dense free-flowing seems to have more distance from the center as compared to both the categories. Reasons being having more number of videos along with having the attention on both the entities as well as the salient regions of the scene. Dense congested has lesser distribution of distance from dense free-flowing but more than sparse because the scene is so congested that the person is unable to focus on something but rather struggles to explore the screen during which the scene changes already.  

\begin{figure}[h!]
\centering
\includegraphics[width=0.4\linewidth]{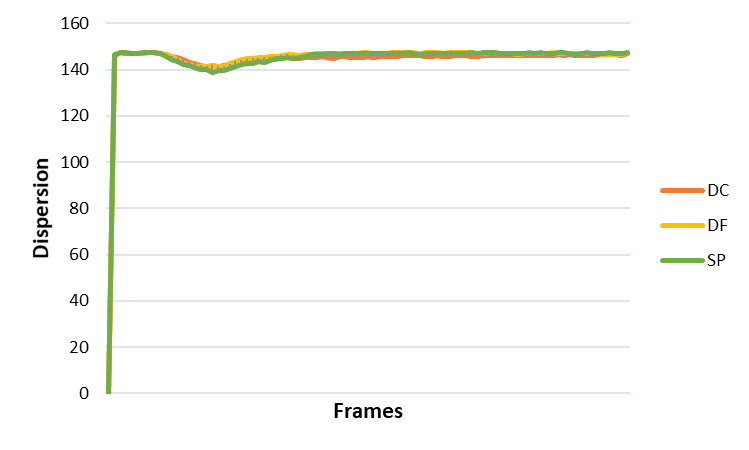}
\caption{Dispersion of Fixations from the Center}
\label{fig:alldisp}
\end{figure}

The spread of the recorded data samples was also evaluated to judge the closeness of agreement between the results. Mostly the standard deviation measure is used for estimating the variability. The fixation coordinates were again used to assess the dispersion of the data around the mean which was later averaged across the participants for all the density levels. Figure~\ref{fig:alldisp} shows the results of different crowd levels.

\newpage

While evaluating the dispersion we can again see the peak at the first frame. The spread remains consistent and around the center showing center-bias. As the attention system lies on perceptual memory, the availability of continuous factors is possible. The small peaks represent the impacts that occurred immediately after jump cuts. 

\subsection{Performance Evaluation over Existing Saliency Models}

Deep Learning models are trained by combining tasks such as feature extraction, integration and saliency value prediction in an end to end manner. Their performance is superior in contrast to classic saliency models. Keeping this in mind, we select two of the latest state-of-the-art dynamic deep learning models~\cite{borji2019saliency}. The models were selected based on best performance on pre-existing dynamic saliency datasets. These models are ACL (resnet variant) ~\cite{wang2018revisiting} and DeepVS ~\cite{jiang2018deepvs}. The third model, SAM ~\cite{cornia2018predicting} is one of the top performing deep static saliency model. 

\begin{table}[!htbp]
\footnotesize
\centering
\begin{tabular*}{450pt}{@{\extracolsep\fill}llcccccccccclD{.}{.}{4}l@{\extracolsep\fill}}
\toprule
Model & Crowd level     & AUC-J   & NSS   & KL     & CC \\ 
\midrule
DeepVS
~ & Sparse              & 0.779	& 0.929	& 1.713 & 0.339   \\
~ & Dense free-flowing  & 0.790	& 1.000	& 1.652 & 0.364   \\
~ & Dense congested     & 0.795	& 1.026	& 1.648 & 0.375   \\
~ & \textbf{Average}    & 0.788	& 0.985	& 1.671 & 0.401  \\
~ & \textit{Baseline}   & 0.90	& 2.94	& 1.24 & 0.57  \\
\midrule
ACL
~ & Sparse              & 0.801	& 1.083	& 1.594 & 0.395  \\
~ & Dense free-flowing  & 0.822	& 1.272	& 1.388 & 0.457  \\
~ & Dense congested     & 0.830	& 1.395	& 1.497 & 0.499  \\
~ & \textbf{Average}    & 0.817	& 1.250	& 1.493 & 0.450  \\
~ & \textit{Baseline}   & 0.890	& 2.354	& --- & 0.434  \\
\midrule
SAM
~ & Sparse              & 0.773	& 0.858	& 1.871 & 0.312 \\ 
~ & Dense free-flowing  & 0.779	& 0.906	& 1.727 & 0.329 \\
~ & Dense congested     & 0.780	& 0.919	& 1.740 & 0.335 \\
~ &\textbf{Average}     & 0.777	& 0.894	& 1.779 & 0.325  \\
~ & \textit{Baseline}   & 0.886	& 3.260	& --- & 0.884  \\
\bottomrule     
\end{tabular*}
\caption{Evaluation Results}
\label{tbl:model_eval}
\end{table}

\begin{figure}[!htbp]
\centering
\subfloat{\includegraphics[width=0.18\linewidth]{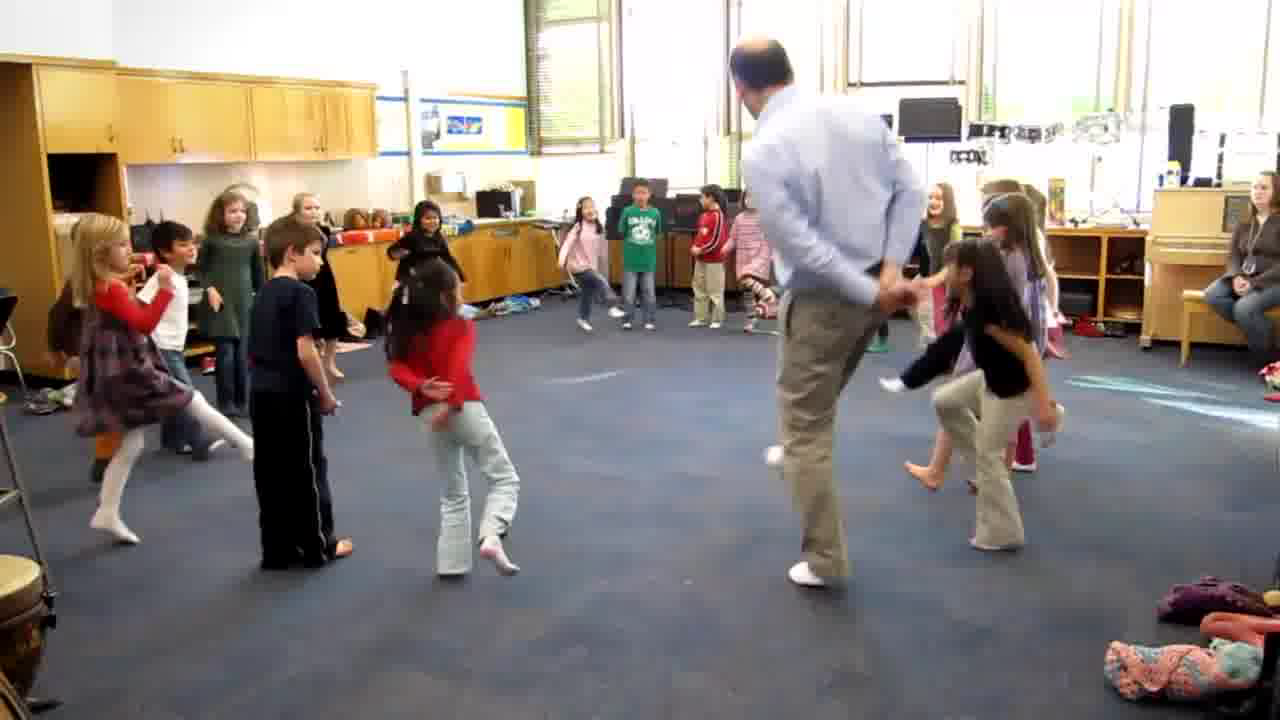}} \quad
\subfloat{\includegraphics[width=0.18\linewidth]{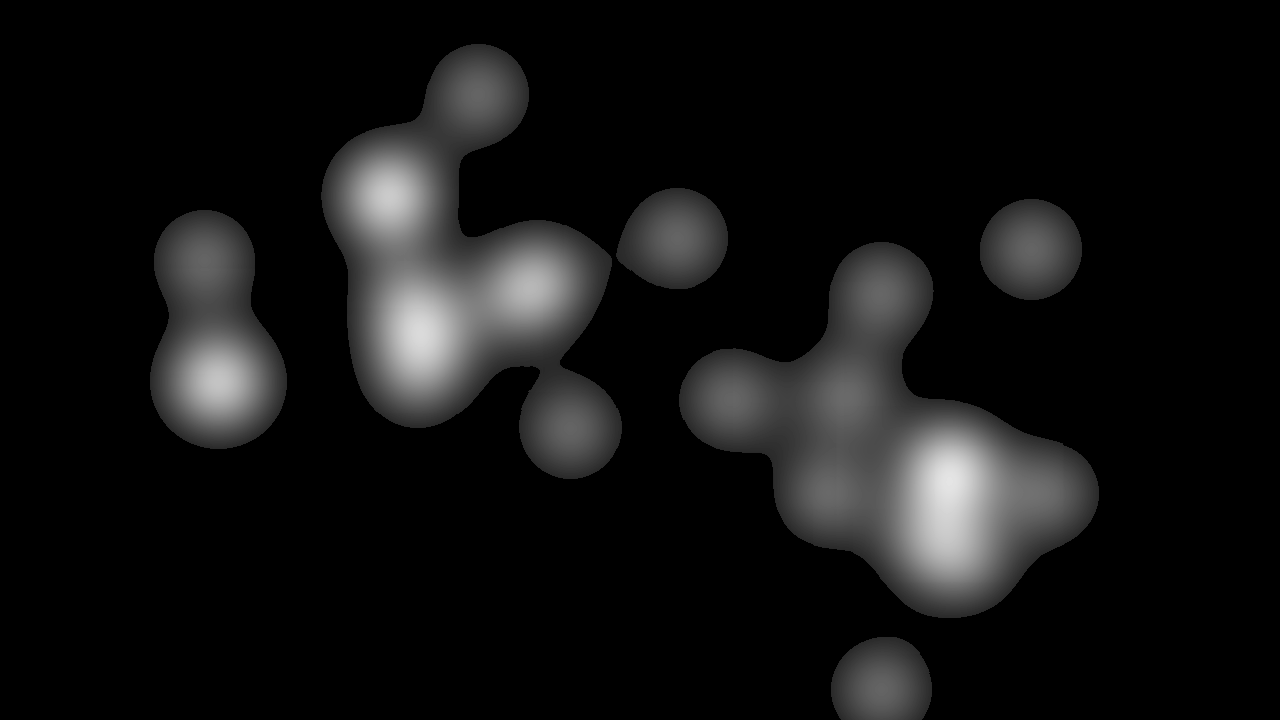}} \quad
\subfloat{\includegraphics[width=0.18\linewidth]{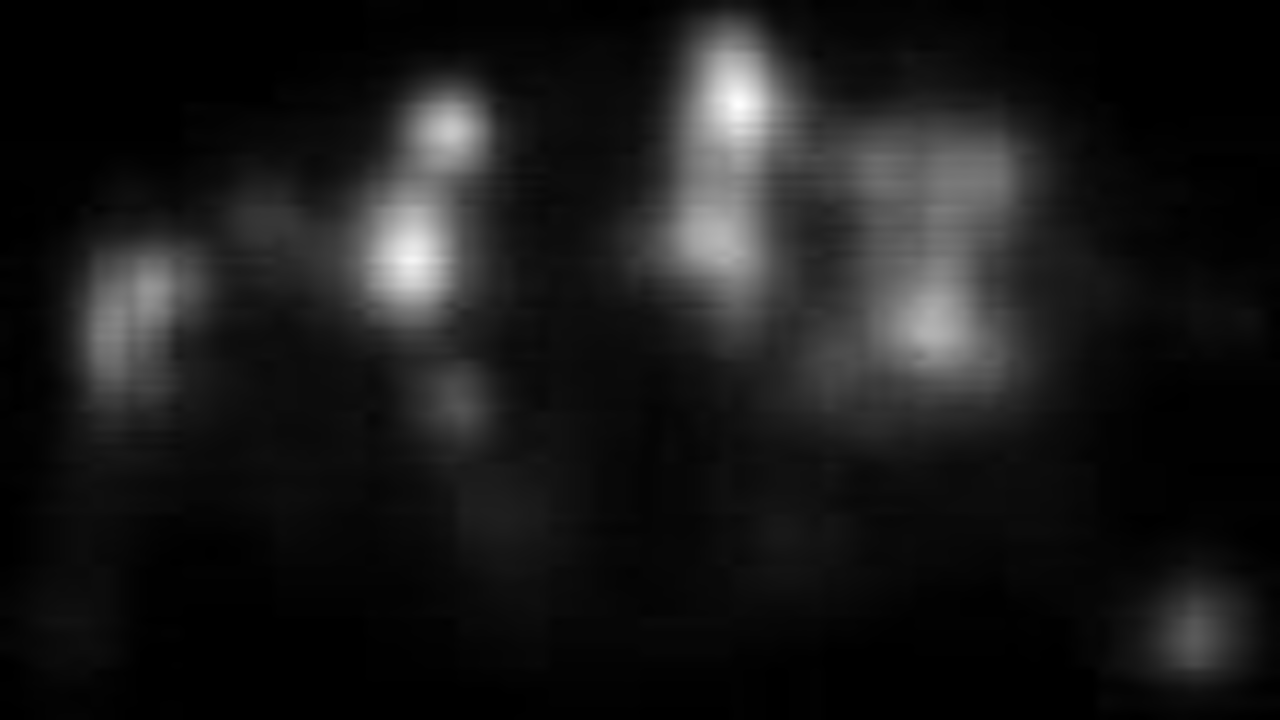}} \quad 
\subfloat{\includegraphics[width=0.18\linewidth]{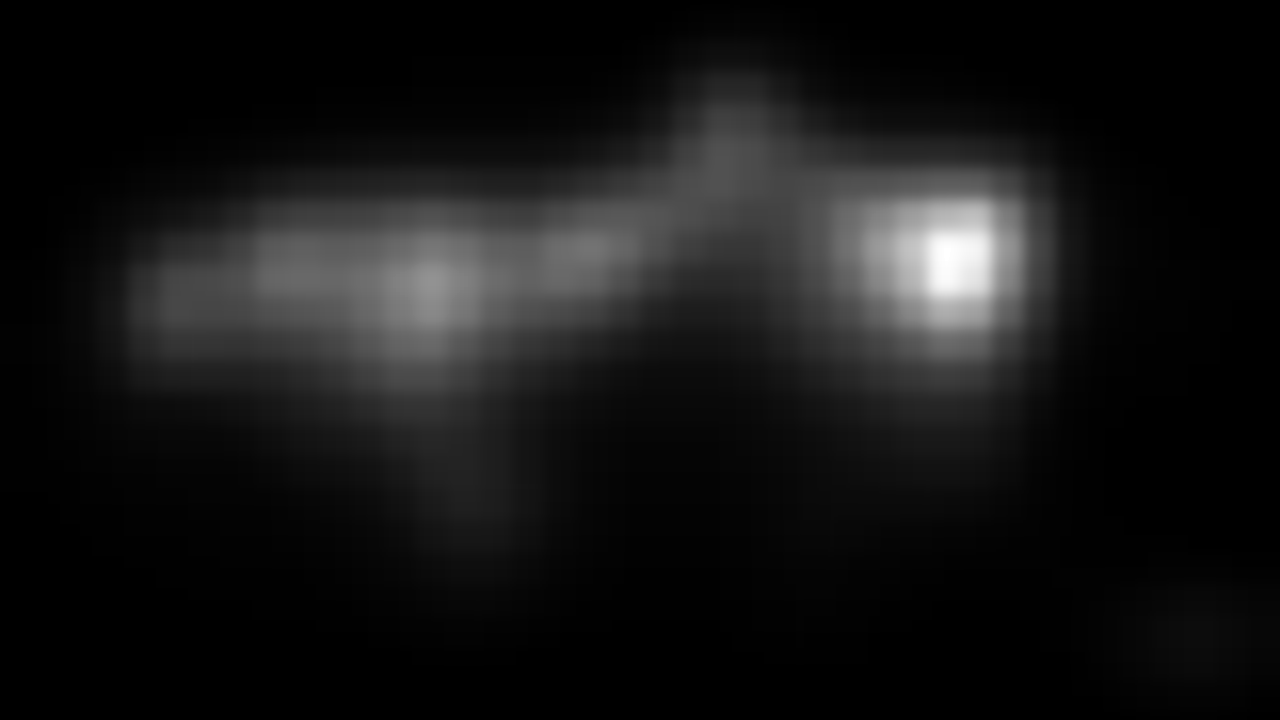}} \quad
\subfloat{\includegraphics[width=0.18\linewidth]{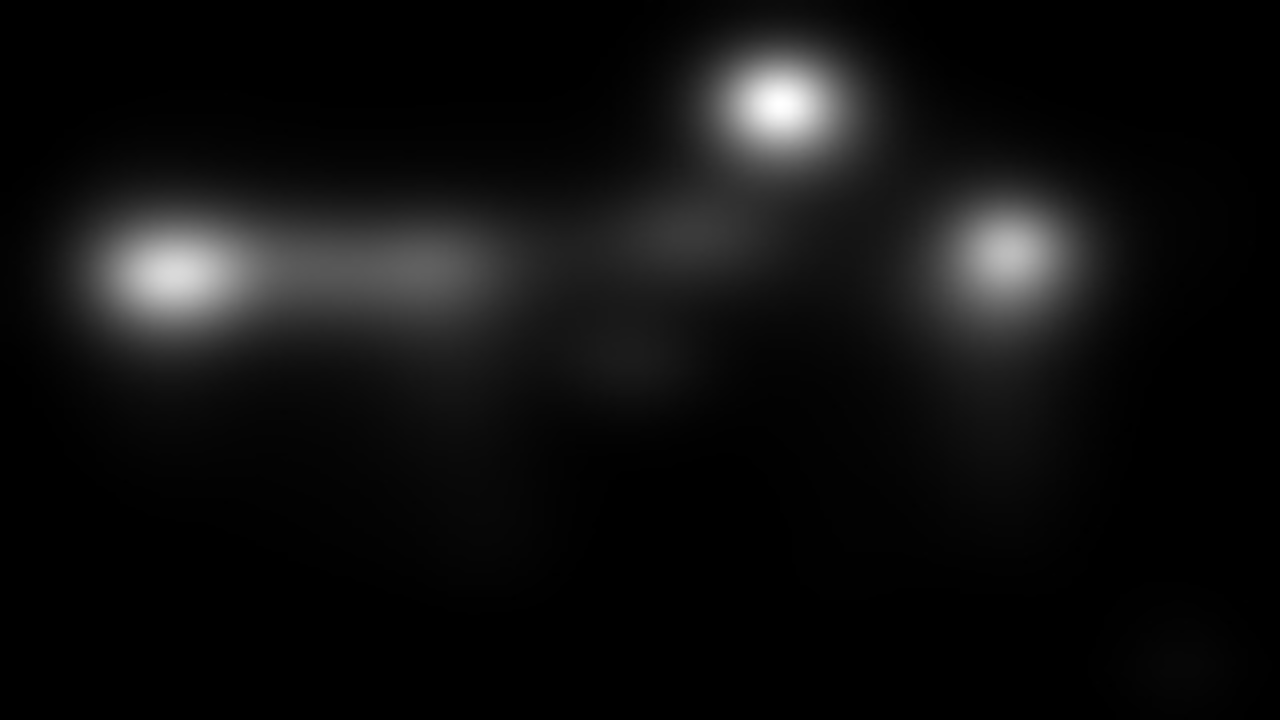}} \\
\subfloat{\includegraphics[width=0.18\linewidth]{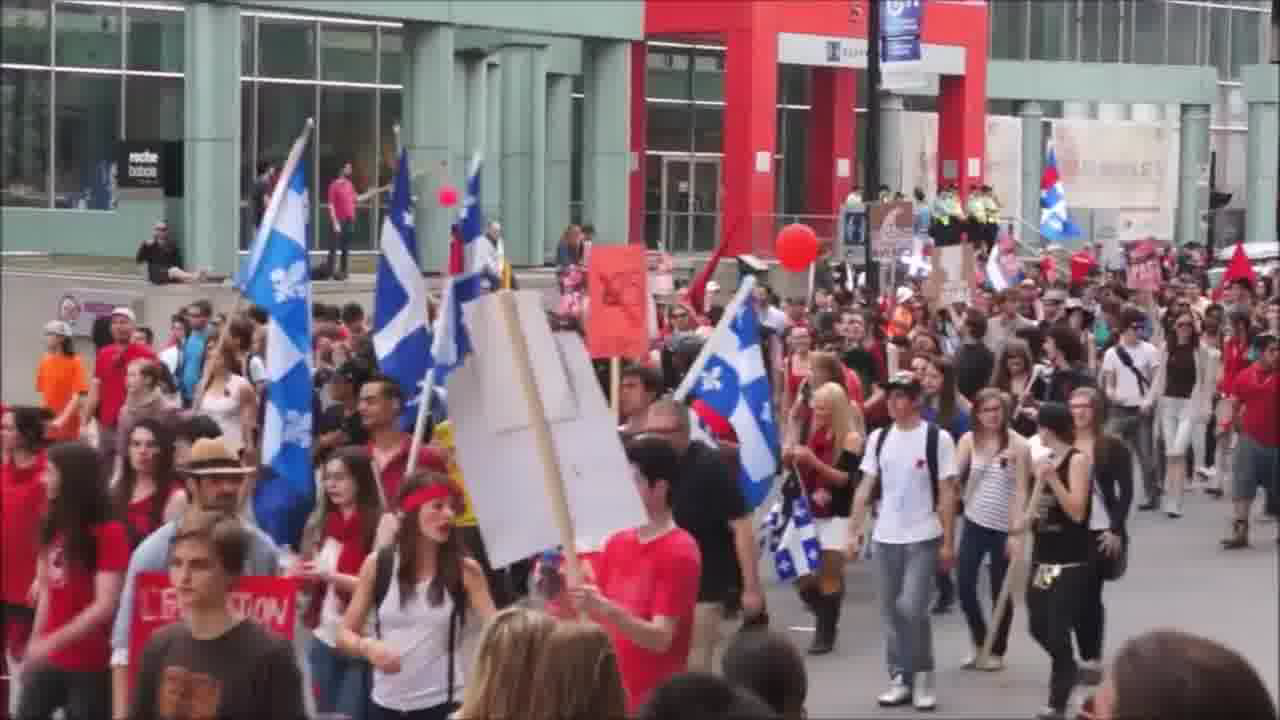}} \quad
\subfloat{\includegraphics[width=0.18\linewidth]{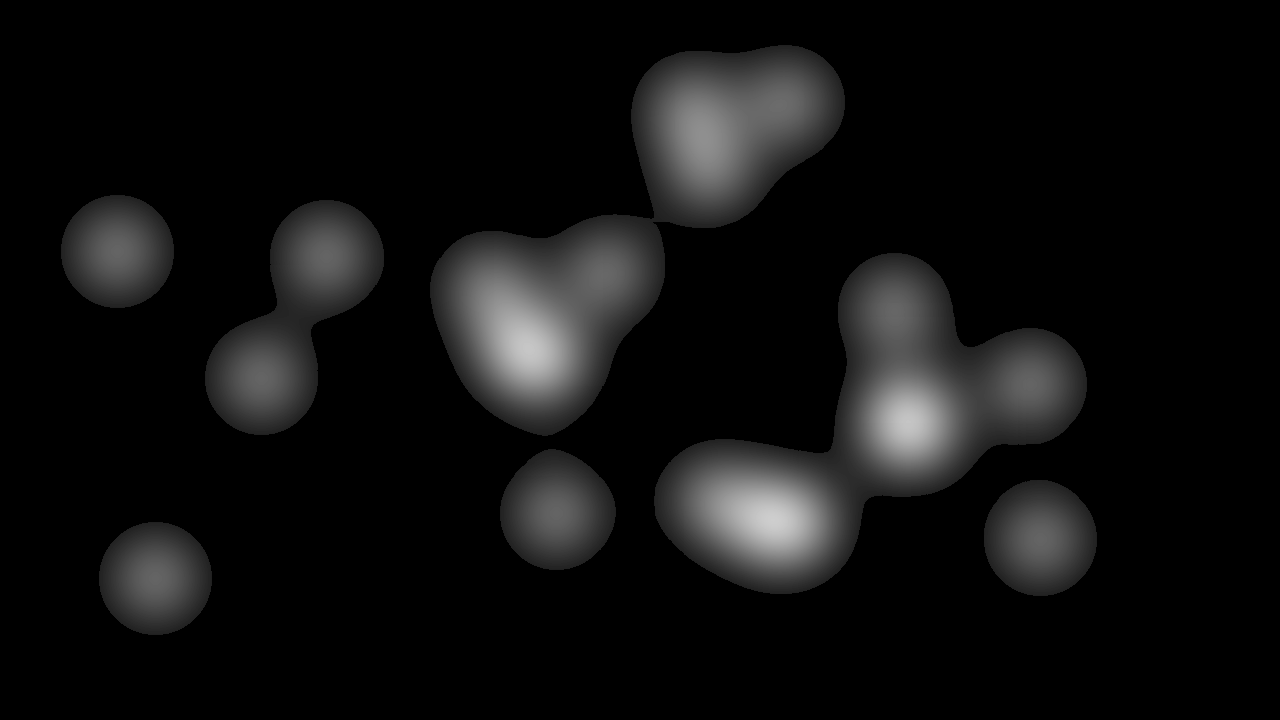}} \quad
\subfloat{\includegraphics[width=0.18\linewidth]{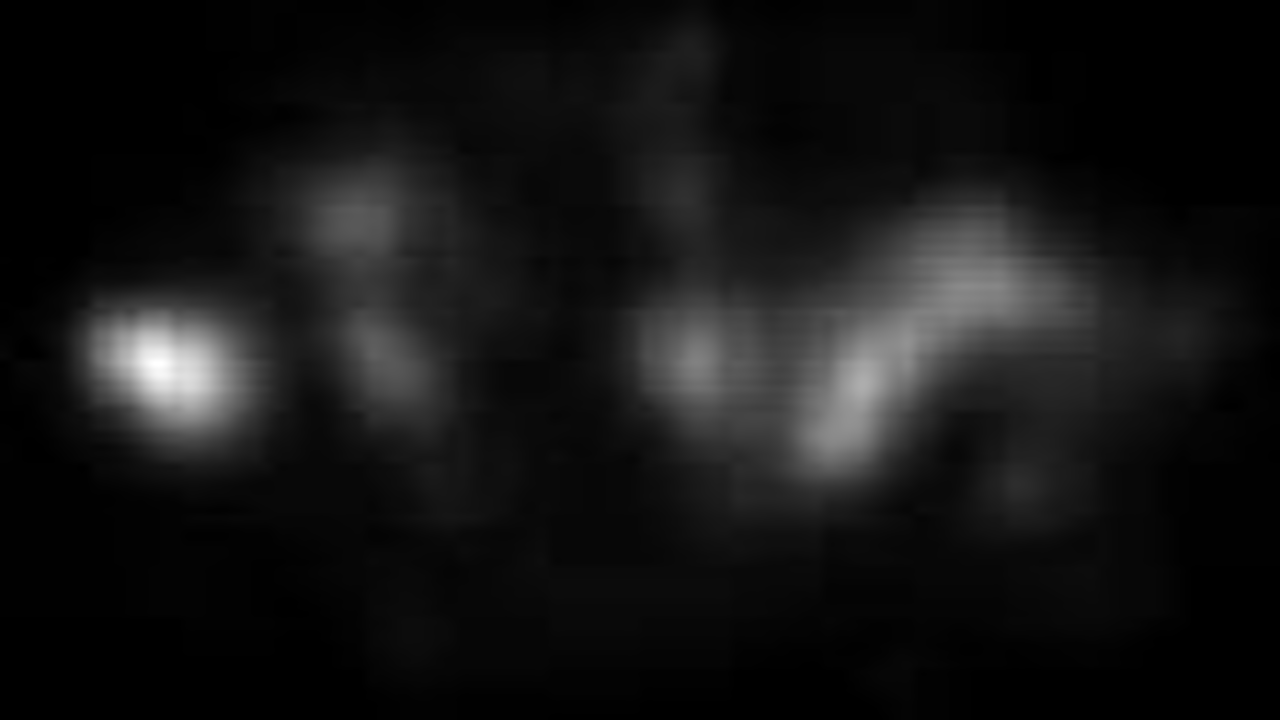}} \quad
\subfloat{\includegraphics[width=0.18\linewidth]{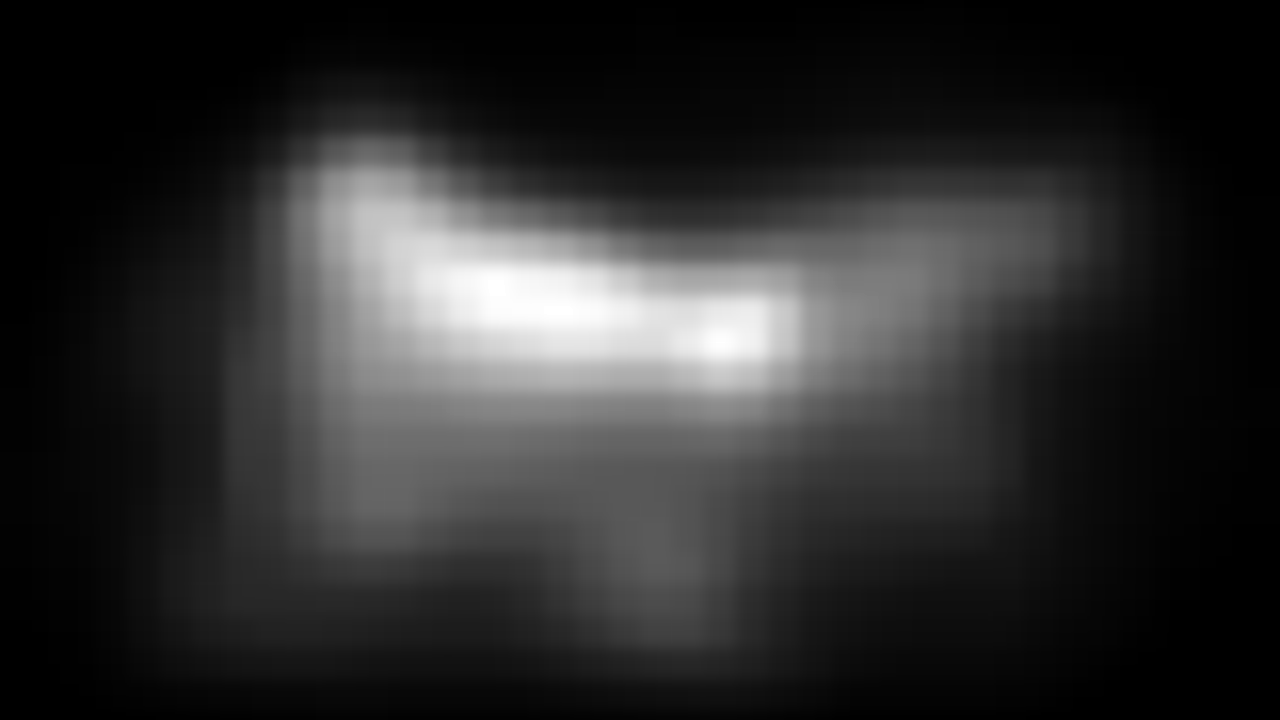}} \quad
\subfloat{\includegraphics[width=0.18\linewidth]{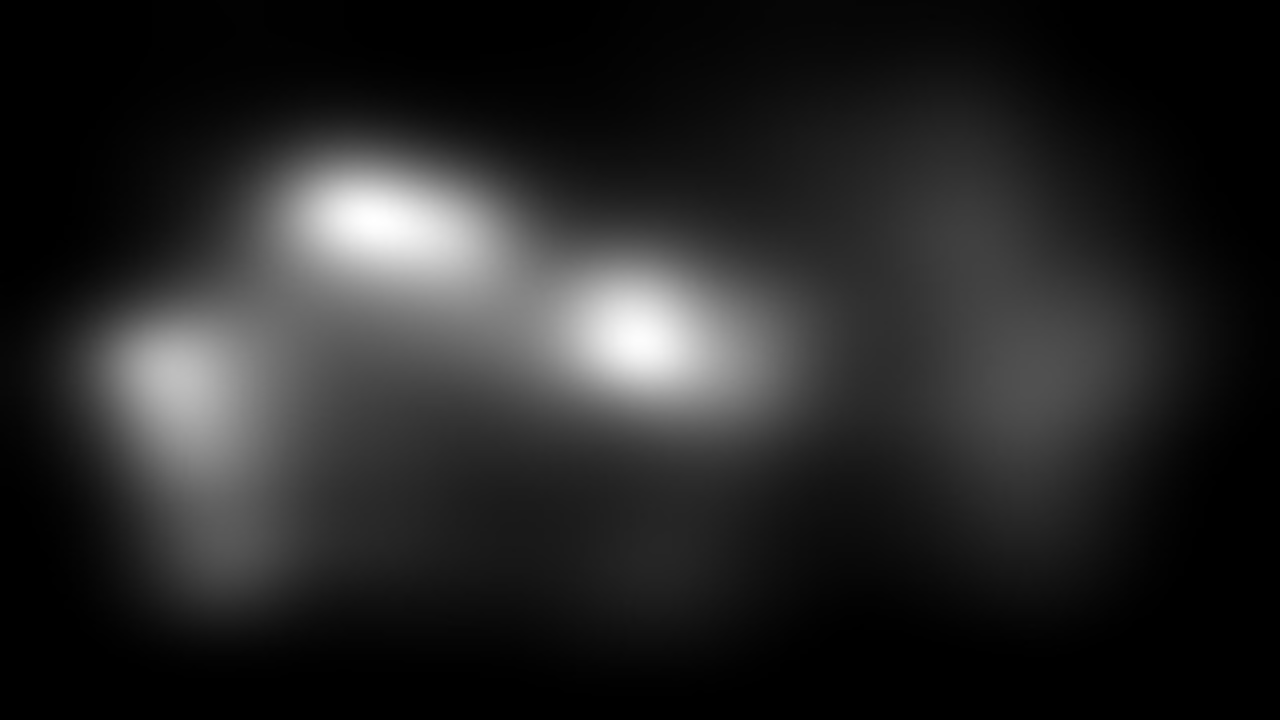}} \\  
\subfloat{\includegraphics[width=0.18\linewidth]{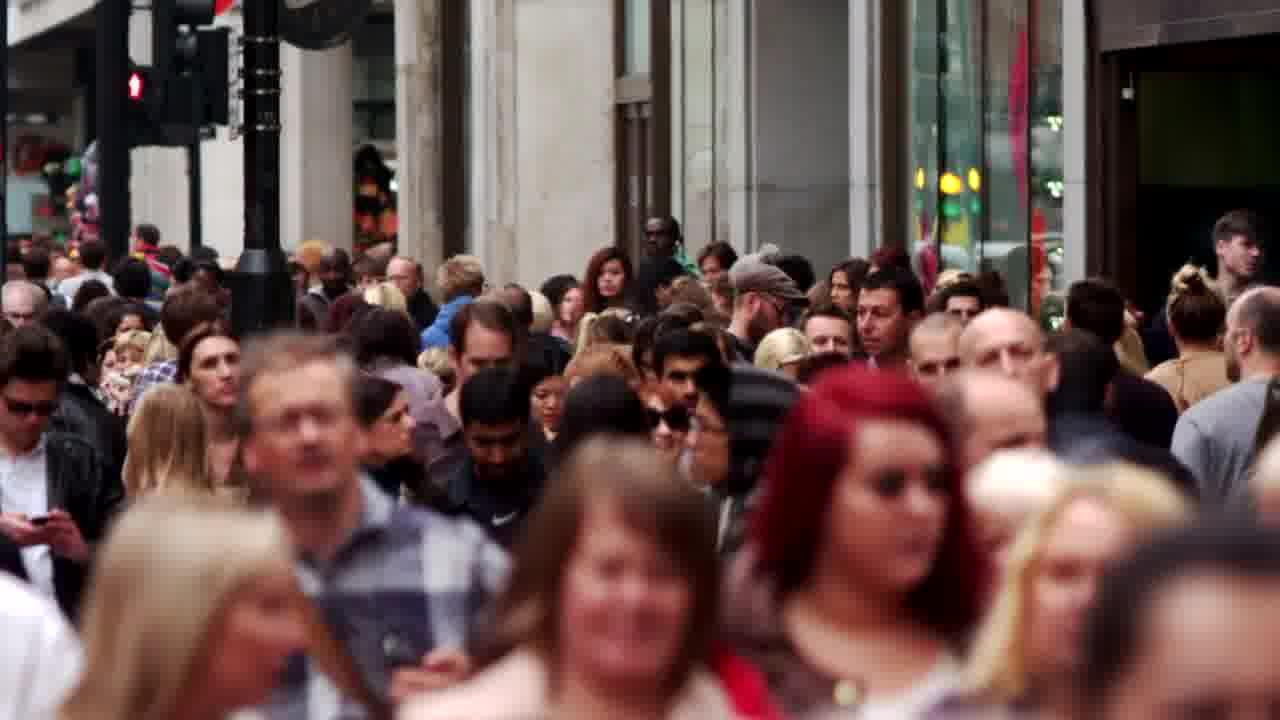}} \quad
\subfloat{\includegraphics[width=0.18\linewidth]{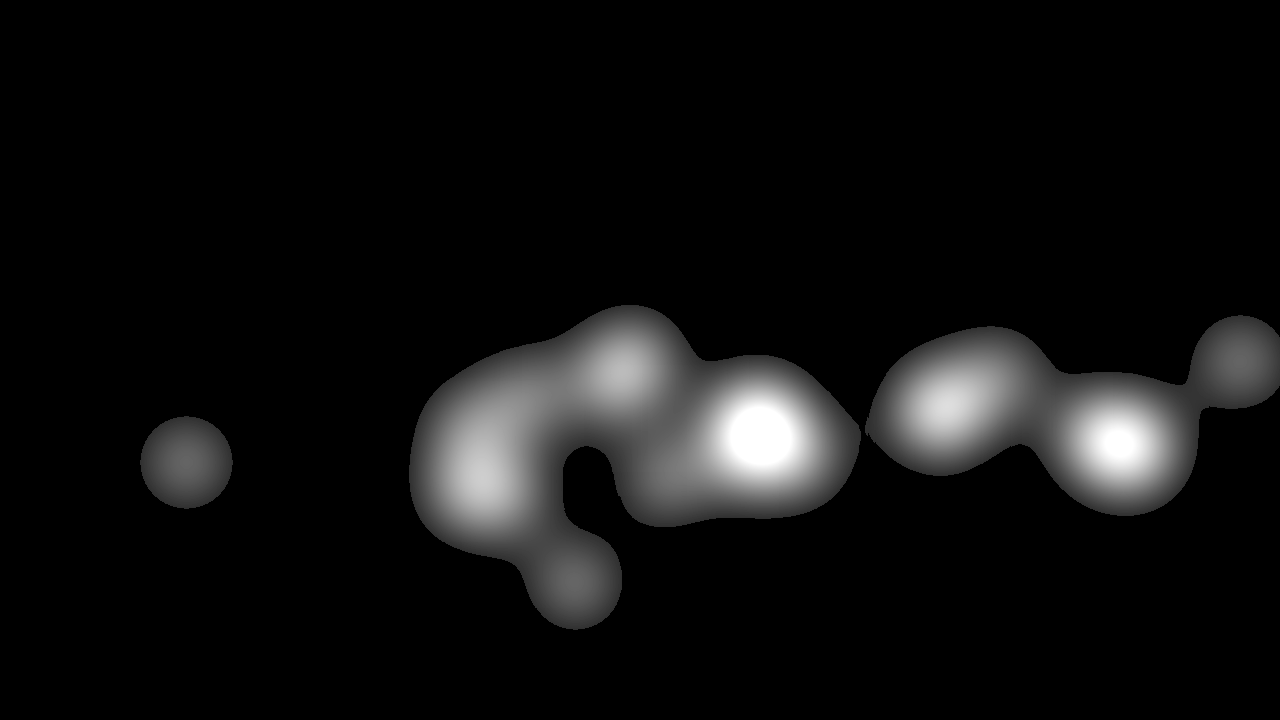}} \quad
\subfloat{\includegraphics[width=0.18\linewidth]{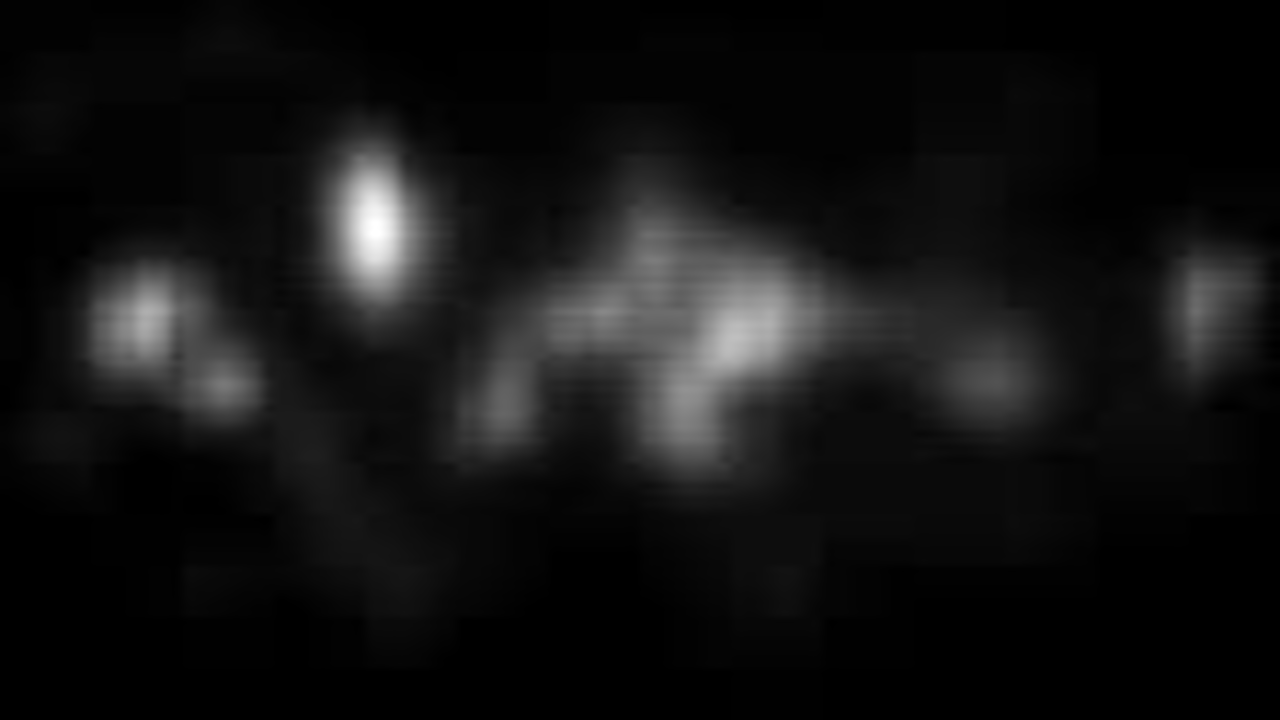}} \quad
\subfloat{\includegraphics[width=0.18\linewidth]{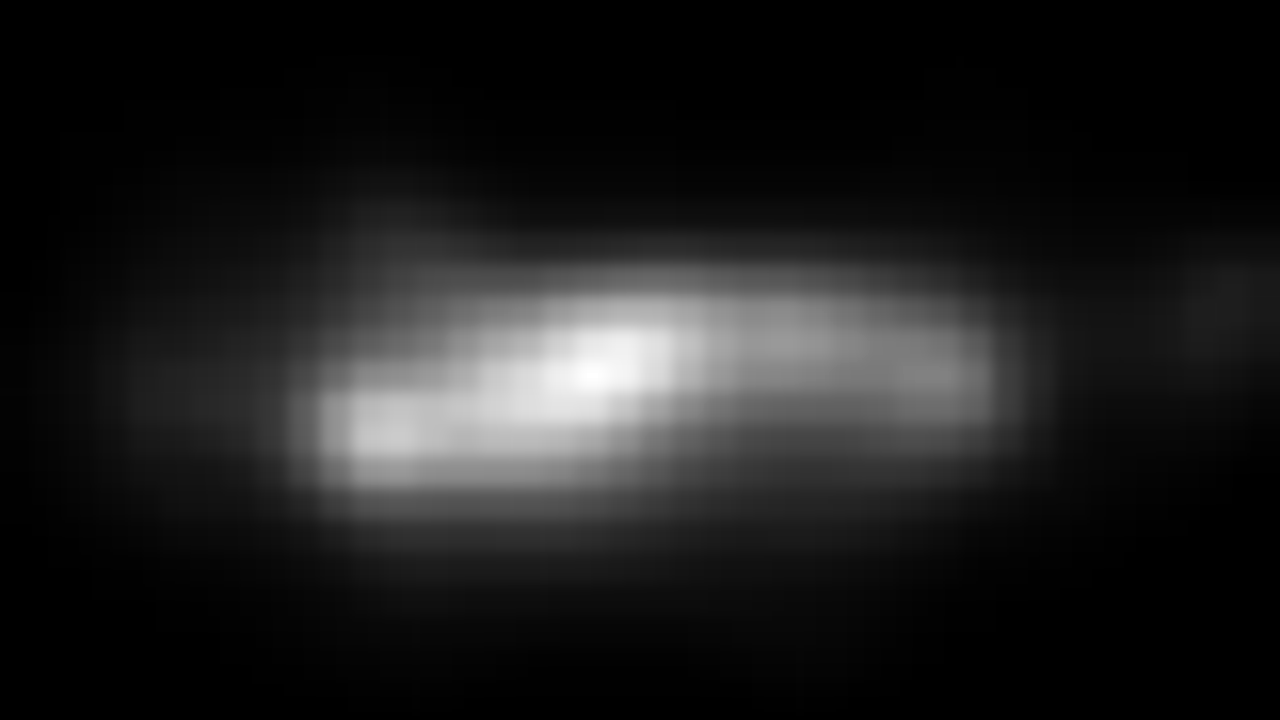}} \quad
\subfloat{\includegraphics[width=0.18\linewidth]{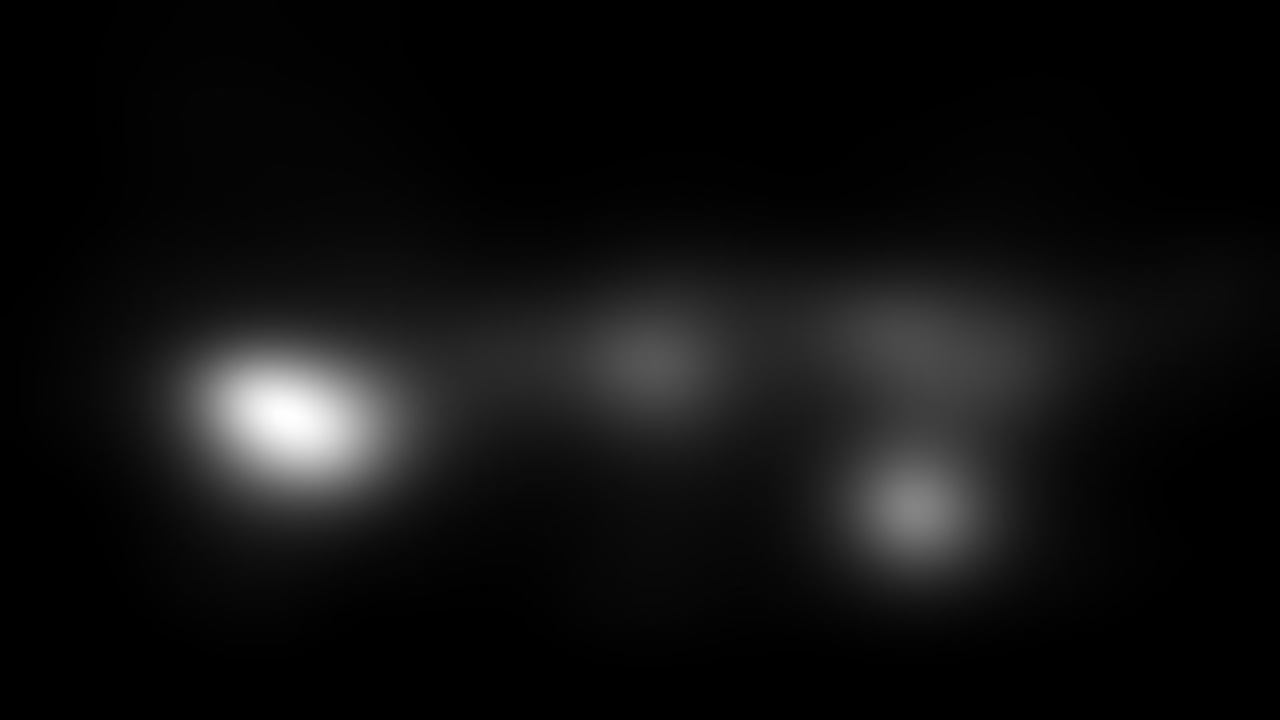}}
\caption{Qualitative results: From left to right, sample video frame, Gaussianed fixation map, and DeepVS, ACL and SAM saliency maps, respectively. From top to bottom: sample video frames for crowd levels sparse, dense free-flowing and dense congested, respectively.}
\label{fig:df}
\end{figure}

We create a benchmark of these models over our dataset. The three models were tested over videos from each crowd category. We choose 4 of the most common saliency evaluation metrics AUC-J, NSS, KLdiv and CC to provide an easy comparison to other saliency benchmarks such as MIT@saliency ~\cite{mit-saliency-benchmark} and the DHF1K video saliency leaderboard. ~\cite{cheng_2019} 
We also provide a baseline from the model's own performance results over their original datasets. We average our results as well to make a comparison with the baseline results and evaluate the performance difference. Figure~\ref{fig:df} shows the original image and its ground truth saliency map for dense free-flowing crowd category. Table~\ref{tbl:model_eval} shows the results of evaluation with DeepVS, ACL and SAM model over different crowd density levels.

Based on the results, ACL performs the best out of three models over all three categories of videos individually and on average. However, the difference between these and ACL's original results is enough to prompt for improvements in model parameter design and architecture to bring saliency prediction in crowds up to par to general saliency prediction. Even in the other two models, the difference between average results and the baseline shows crowd videos need customized saliency prediction models to reach state-of-art-performance.

\section{Discussion and Conclusion}

Crowd Scenes provide a richer set of dynamics and stimuli. These can be used to formulate and test the accuracy of general saliency judgments and models if they hold true fro crowd scenes as well provide insights on how to bring about improvements.

In this work, we studied the crowd characteristics and categorised the crowds into different density levels. The fixation and dispersion analysis shows that attention does vary with the number of people in the crowd. As the crowd gets bigger, most of the time is spent viewing more objects in the scene rather than paying attention to any one particular object. With decrease in the number of entities, salient features are more spontaneously noticeable in individual objects. As s future avenue, to bridge the gap in human performance and predicted saliency, it would be prudent to include more cognitive information about crowded stimuli into the computational models~\cite{feng2016fixation}. The importance of different features particularly facial features in the context of crowd videos is still an unexplored area. General saliency datasets' evaluation metrics results and ours reflect a quite a big gap in performance. There is an obvious gap for improving deep saliency model to work equally well, if not better for crowds. We reiterate the need to investigate which features should be reinforced for crowd videos in the design of the model to predict better crowd scene saliency. 

\bibliography{references}

\begin{thebibliography}{}

\bibitem[\protect\citeauthoryear{%
Blignaut%
\ \BBA{} Wium%
}{%
Blignaut%
\ \BBA{} Wium%
}{%
{\protect\APACyear{2014}}%
}]{%
blignaut2014eye}%
\APACinsertmetastar{%
blignaut2014eye}%
Blignaut, P.%
\BCBT{}\ \BBA{} Wium, D.%
%
\unskip\
\newblock
\APACrefYearMonthDay{2014}{}{}.
\newblock
\BBOQ{}\APACrefatitle{Eye-tracking data quality as affected by ethnicity and
  experimental design}{Eye-tracking data quality as affected by ethnicity and
  experimental design}.\BBCQ{}
\newblock
\APACjournalVolNumPages{Behavior research methods}{46}{1}{67--80}.
\PrintBackRefs{\CurrentBib}

\bibitem[\protect\citeauthoryear{%
Borji%
}{%
Borji%
}{%
{\protect\APACyear{2019}}%
}]{%
borji2019saliency}%
\APACinsertmetastar{%
borji2019saliency}%
Borji, A.%
%
\unskip\
\newblock
\APACrefYearMonthDay{2019}{}{}.
\newblock
\BBOQ{}\APACrefatitle{Saliency Prediction in the Deep Learning Era: Successes
  and Limitations}{Saliency prediction in the deep learning era: Successes and
  limitations}.\BBCQ{}
\newblock
\APACjournalVolNumPages{IEEE transactions on pattern analysis and machine
  intelligence}{}{}{}.
\PrintBackRefs{\CurrentBib}

\bibitem[\protect\citeauthoryear{%
Borji%
\ \BBA{} Itti%
}{%
Borji%
\ \BBA{} Itti%
}{%
{\protect\APACyear{2012}}%
}]{%
borji2012state}%
\APACinsertmetastar{%
borji2012state}%
Borji, A.%
\BCBT{}\ \BBA{} Itti, L.%
%
\unskip\
\newblock
\APACrefYearMonthDay{2012}{}{}.
\newblock
\BBOQ{}\APACrefatitle{State-of-the-art in visual attention
  modeling}{State-of-the-art in visual attention modeling}.\BBCQ{}
\newblock
\APACjournalVolNumPages{IEEE transactions on pattern analysis and machine
  intelligence}{35}{1}{185--207}.
\PrintBackRefs{\CurrentBib}

\bibitem[\protect\citeauthoryear{%
Bylinskii%
\ \protect\BOthers{.}}{%
Bylinskii%
\ \protect\BOthers{.}}{%
{\protect\APACyear{2019}}%
}]{%
mit-saliency-benchmark}%
\APACinsertmetastar{%
mit-saliency-benchmark}%
Bylinskii, Z.%
, Judd, T.%
, Borji, A.%
, Itti, L.%
, Durand, F.%
, Oliva, A.%
\BCBL{}\ \BOthersPeriod{.}%
\unskip\
\newblock
\APACrefYearMonthDay{2019}{Sep}{}.
\newblock
\APACrefbtitle{MIT Saliency Benchmark.}{Mit saliency benchmark.}
\newblock
 \begin{APACrefURL} \url{http://saliency.mit.edu/results_mit300.html}
  \end{APACrefURL}
\PrintBackRefs{\CurrentBib}

\bibitem[\protect\citeauthoryear{%
Carmi%
\ \BBA{} Itti%
}{%
Carmi%
\ \BBA{} Itti%
}{%
{\protect\APACyear{2006}}%
}]{%
carmi2006role}%
\APACinsertmetastar{%
carmi2006role}%
Carmi, R.%
\BCBT{}\ \BBA{} Itti, L.%
%
\unskip\
\newblock
\APACrefYearMonthDay{2006}{}{}.
\newblock
\BBOQ{}\APACrefatitle{The role of memory in guiding attention during natural
  vision}{The role of memory in guiding attention during natural
  vision}.\BBCQ{}
\newblock
\APACjournalVolNumPages{Journal of vision}{6}{9}{4--4}.
\PrintBackRefs{\CurrentBib}

\bibitem[\protect\citeauthoryear{%
Cheng%
}{%
Cheng%
}{%
{\protect\APACyear{2019}}%
}]{%
cheng_2019}%
\APACinsertmetastar{%
cheng_2019}%
Cheng, M\BHBI M.%
%
\unskip\
\newblock
\APACrefYearMonthDay{2019}{Sep}{}.
\newblock
\APACrefbtitle{Revisiting Video Saliency Prediction in the Deep Learning
  Era.}{Revisiting video saliency prediction in the deep learning era.}
\newblock
 \begin{APACrefURL} \url{https://mmcheng.net/videosal/} \end{APACrefURL}
\PrintBackRefs{\CurrentBib}

\bibitem[\protect\citeauthoryear{%
Chiappino%
, Morerio%
, Marcenaro%
\BCBL{}\ \BBA{} Regazzoni%
}{%
Chiappino%
\ \protect\BOthers{.}}{%
{\protect\APACyear{2015}}%
}]{%
chiappino2015bio}%
\APACinsertmetastar{%
chiappino2015bio}%
Chiappino, S.%
, Morerio, P.%
, Marcenaro, L.%
\BCBL{}\ \BBA{} Regazzoni, C\BPBI S.%
%
\unskip\
\newblock
\APACrefYearMonthDay{2015}{}{}.
\newblock
\BBOQ{}\APACrefatitle{Bio-inspired relevant interaction modelling in cognitive
  crowd management}{Bio-inspired relevant interaction modelling in cognitive
  crowd management}.\BBCQ{}
\newblock
\APACjournalVolNumPages{Journal of Ambient Intelligence and Humanized
  Computing}{6}{2}{171--192}.
\PrintBackRefs{\CurrentBib}

\bibitem[\protect\citeauthoryear{%
Cornia%
, Baraldi%
, Serra%
\BCBL{}\ \BBA{} Cucchiara%
}{%
Cornia%
\ \protect\BOthers{.}}{%
{\protect\APACyear{2018}}%
}]{%
cornia2018predicting}%
\APACinsertmetastar{%
cornia2018predicting}%
Cornia, M.%
, Baraldi, L.%
, Serra, G.%
\BCBL{}\ \BBA{} Cucchiara, R.%
%
\unskip\
\newblock
\APACrefYearMonthDay{2018}{}{}.
\newblock
\BBOQ{}\APACrefatitle{Predicting human eye fixations via an lstm-based saliency
  attentive model}{Predicting human eye fixations via an lstm-based saliency
  attentive model}.\BBCQ{}
\newblock
\APACjournalVolNumPages{IEEE Transactions on Image
  Processing}{27}{10}{5142--5154}.
\PrintBackRefs{\CurrentBib}

\bibitem[\protect\citeauthoryear{%
Dalrymple%
, Manner%
, Harmelink%
, Teska%
\BCBL{}\ \BBA{} Elison%
}{%
Dalrymple%
\ \protect\BOthers{.}}{%
{\protect\APACyear{2018}}%
}]{%
dalrymple2018examination}%
\APACinsertmetastar{%
dalrymple2018examination}%
Dalrymple, K\BPBI A.%
, Manner, M\BPBI D.%
, Harmelink, K\BPBI A.%
, Teska, E\BPBI P.%
\BCBL{}\ \BBA{} Elison, J\BPBI T.%
%
\unskip\
\newblock
\APACrefYearMonthDay{2018}{}{}.
\newblock
\BBOQ{}\APACrefatitle{An examination of recording accuracy and precision from
  eye tracking data from toddlerhood to adulthood}{An examination of recording
  accuracy and precision from eye tracking data from toddlerhood to
  adulthood}.\BBCQ{}
\newblock
\APACjournalVolNumPages{Frontiers in psychology}{9}{}{}.
\PrintBackRefs{\CurrentBib}

\bibitem[\protect\citeauthoryear{%
Feng%
, Borji%
\BCBL{}\ \BBA{} Lu%
}{%
Feng%
\ \protect\BOthers{.}}{%
{\protect\APACyear{2016}}%
}]{%
feng2016fixation}%
\APACinsertmetastar{%
feng2016fixation}%
Feng, M.%
, Borji, A.%
\BCBL{}\ \BBA{} Lu, H.%
%
\unskip\
\newblock
\APACrefYearMonthDay{2016}{}{}.
\newblock
\BBOQ{}\APACrefatitle{Fixation prediction with a combined model of bottom-up
  saliency and vanishing point}{Fixation prediction with a combined model of
  bottom-up saliency and vanishing point}.\BBCQ{}
\newblock
\BIn{} \APACrefbtitle{2016 IEEE Winter Conference on Applications of Computer
  Vision (WACV)}{2016 ieee winter conference on applications of computer vision
  (wacv)}\ (\BPGS\ 1--7).
\PrintBackRefs{\CurrentBib}

\bibitem[\protect\citeauthoryear{%
Gupta%
\ \BBA{} Gupta%
}{%
Gupta%
\ \BBA{} Gupta%
}{%
{\protect\APACyear{2014}}%
}]{%
gupta2014design}%
\APACinsertmetastar{%
gupta2014design}%
Gupta, J\BPBI K.%
\BCBT{}\ \BBA{} Gupta, S.%
%
\unskip\
\newblock
\APACrefYearMonthDay{2014}{}{}.
\newblock
\BBOQ{}\APACrefatitle{Design and Analysis of Crowd Sized Estimation
  Techniques}{Design and analysis of crowd sized estimation techniques}.\BBCQ{}
\newblock
\APACjournalVolNumPages{International Journal of Computer
  Applications}{107}{19}{}.
\PrintBackRefs{\CurrentBib}

\bibitem[\protect\citeauthoryear{%
He%
, Tavakoli%
, Borji%
, Mi%
\BCBL{}\ \BBA{} Pugeault%
}{%
He%
\ \protect\BOthers{.}}{%
{\protect\APACyear{2019}}%
}]{%
he2019understanding}%
\APACinsertmetastar{%
he2019understanding}%
He, S.%
, Tavakoli, H\BPBI R.%
, Borji, A.%
, Mi, Y.%
\BCBL{}\ \BBA{} Pugeault, N.%
%
\unskip\
\newblock
\APACrefYearMonthDay{2019}{}{}.
\newblock
\BBOQ{}\APACrefatitle{Understanding and Visualizing Deep Visual Saliency
  Models}{Understanding and visualizing deep visual saliency models}.\BBCQ{}
\newblock
\BIn{} \APACrefbtitle{Proceedings of the IEEE Conference on Computer Vision and
  Pattern Recognition}{Proceedings of the ieee conference on computer vision
  and pattern recognition}\ (\BPGS\ 10206--10215).
\PrintBackRefs{\CurrentBib}

\bibitem[\protect\citeauthoryear{%
Holmqvist%
, Nystr{\"o}m%
\BCBL{}\ \BBA{} Mulvey%
}{%
Holmqvist%
\ \protect\BOthers{.}}{%
{\protect\APACyear{2012}}%
}]{%
holmqvist2012eye}%
\APACinsertmetastar{%
holmqvist2012eye}%
Holmqvist, K.%
, Nystr{\"o}m, M.%
\BCBL{}\ \BBA{} Mulvey, F.%
%
\unskip\
\newblock
\APACrefYearMonthDay{2012}{}{}.
\newblock
\BBOQ{}\APACrefatitle{Eye tracker data quality: what it is and how to measure
  it}{Eye tracker data quality: what it is and how to measure it}.\BBCQ{}
\newblock
\BIn{} \APACrefbtitle{Proceedings of the symposium on eye tracking research and
  applications}{Proceedings of the symposium on eye tracking research and
  applications}\ (\BPGS\ 45--52).
\PrintBackRefs{\CurrentBib}

\bibitem[\protect\citeauthoryear{%
L.~Jiang%
, Xu%
, Liu%
, Qiao%
\BCBL{}\ \BBA{} Wang%
}{%
L.~Jiang%
\ \protect\BOthers{.}}{%
{\protect\APACyear{2018}}%
}]{%
jiang2018deepvs}%
\APACinsertmetastar{%
jiang2018deepvs}%
Jiang, L.%
, Xu, M.%
, Liu, T.%
, Qiao, M.%
\BCBL{}\ \BBA{} Wang, Z.%
%
\unskip\
\newblock
\APACrefYearMonthDay{2018}{}{}.
\newblock
\BBOQ{}\APACrefatitle{Deepvs: A deep learning based video saliency prediction
  approach}{Deepvs: A deep learning based video saliency prediction
  approach}.\BBCQ{}
\newblock
\BIn{} \APACrefbtitle{Proceedings of the European Conference on Computer Vision
  (ECCV)}{Proceedings of the european conference on computer vision (eccv)}\
  (\BPGS\ 602--617).
\PrintBackRefs{\CurrentBib}

\bibitem[\protect\citeauthoryear{%
M.~Jiang%
, Xu%
\BCBL{}\ \BBA{} Zhao%
}{%
M.~Jiang%
\ \protect\BOthers{.}}{%
{\protect\APACyear{2014}}%
}]{%
jiang2014people}%
\APACinsertmetastar{%
jiang2014people}%
Jiang, M.%
, Xu, J.%
\BCBL{}\ \BBA{} Zhao, Q.%
%
\unskip\
\newblock
\APACrefYearMonthDay{2014}{}{}.
\newblock
\BBOQ{}\APACrefatitle{Where Do People Look at in Crowded Natural Scenes?}{Where
  do people look at in crowded natural scenes?}\BBCQ{}
\newblock
\APACjournalVolNumPages{Journal of Vision}{14}{10}{1052--1052}.
\PrintBackRefs{\CurrentBib}

\bibitem[\protect\citeauthoryear{%
Judd%
, Ehinger%
, Durand%
\BCBL{}\ \BBA{} Torralba%
}{%
Judd%
\ \protect\BOthers{.}}{%
{\protect\APACyear{2009}}%
}]{%
judd2009learning}%
\APACinsertmetastar{%
judd2009learning}%
Judd, T.%
, Ehinger, K.%
, Durand, F.%
\BCBL{}\ \BBA{} Torralba, A.%
%
\unskip\
\newblock
\APACrefYearMonthDay{2009}{}{}.
\newblock
\BBOQ{}\APACrefatitle{Learning to predict where humans look}{Learning to
  predict where humans look}.\BBCQ{}
\newblock
\BIn{} \APACrefbtitle{2009 IEEE 12th international conference on computer
  vision}{2009 ieee 12th international conference on computer vision}\ (\BPGS\
  2106--2113).
\PrintBackRefs{\CurrentBib}

\bibitem[\protect\citeauthoryear{%
Le~Meur%
, Le~Callet%
, Barba%
\BCBL{}\ \BBA{} Thoreau%
}{%
Le~Meur%
\ \protect\BOthers{.}}{%
{\protect\APACyear{2006}}%
}]{%
le2006coherent}%
\APACinsertmetastar{%
le2006coherent}%
Le~Meur, O.%
, Le~Callet, P.%
, Barba, D.%
\BCBL{}\ \BBA{} Thoreau, D.%
%
\unskip\
\newblock
\APACrefYearMonthDay{2006}{}{}.
\newblock
\BBOQ{}\APACrefatitle{A coherent computational approach to model bottom-up
  visual attention}{A coherent computational approach to model bottom-up visual
  attention}.\BBCQ{}
\newblock
\APACjournalVolNumPages{IEEE transactions on pattern analysis and machine
  intelligence}{28}{5}{802--817}.
\PrintBackRefs{\CurrentBib}

\bibitem[\protect\citeauthoryear{%
Mancas%
}{%
Mancas%
}{%
{\protect\APACyear{2010}}%
}]{%
mancas2010attention}%
\APACinsertmetastar{%
mancas2010attention}%
Mancas, M.%
%
\unskip\
\newblock
\APACrefYearMonthDay{2010}{}{}.
\newblock
\BBOQ{}\APACrefatitle{Attention-based dense crowds analysis}{Attention-based
  dense crowds analysis}.\BBCQ{}
\newblock
\BIn{} \APACrefbtitle{11th International Workshop on Image Analysis for
  Multimedia Interactive Services WIAMIS 10}{11th international workshop on
  image analysis for multimedia interactive services wiamis 10}\ (\BPGS\ 1--4).
\PrintBackRefs{\CurrentBib}

\bibitem[\protect\citeauthoryear{%
Mancas%
\ \BBA{} Gosselin%
}{%
Mancas%
\ \BBA{} Gosselin%
}{%
{\protect\APACyear{2010}}%
}]{%
mancas2010dense}%
\APACinsertmetastar{%
mancas2010dense}%
Mancas, M.%
\BCBT{}\ \BBA{} Gosselin, B.%
%
\unskip\
\newblock
\APACrefYearMonthDay{2010}{}{}.
\newblock
\BBOQ{}\APACrefatitle{Dense crowd analysis through bottom-up and top-down
  attention}{Dense crowd analysis through bottom-up and top-down
  attention}.\BBCQ{}
\newblock
\BIn{} \APACrefbtitle{Proc. Brain Inspired Cognit. Syst.(BICS)}{Proc. brain
  inspired cognit. syst.(bics)}\ (\BPGS\ 1--12).
\PrintBackRefs{\CurrentBib}

\bibitem[\protect\citeauthoryear{%
Mathe%
\ \BBA{} Sminchisescu%
}{%
Mathe%
\ \BBA{} Sminchisescu%
}{%
{\protect\APACyear{2014}}%
}]{%
mathe2014actions}%
\APACinsertmetastar{%
mathe2014actions}%
Mathe, S.%
\BCBT{}\ \BBA{} Sminchisescu, C.%
%
\unskip\
\newblock
\APACrefYearMonthDay{2014}{}{}.
\newblock
\BBOQ{}\APACrefatitle{Actions in the eye: Dynamic gaze datasets and learnt
  saliency models for visual recognition}{Actions in the eye: Dynamic gaze
  datasets and learnt saliency models for visual recognition}.\BBCQ{}
\newblock
\APACjournalVolNumPages{IEEE transactions on pattern analysis and machine
  intelligence}{37}{7}{1408--1424}.
\PrintBackRefs{\CurrentBib}

\bibitem[\protect\citeauthoryear{%
Mital%
, Smith%
, Hill%
\BCBL{}\ \BBA{} Henderson%
}{%
Mital%
\ \protect\BOthers{.}}{%
{\protect\APACyear{2011}}%
}]{%
mital2011clustering}%
\APACinsertmetastar{%
mital2011clustering}%
Mital, P\BPBI K.%
, Smith, T\BPBI J.%
, Hill, R\BPBI L.%
\BCBL{}\ \BBA{} Henderson, J\BPBI M.%
%
\unskip\
\newblock
\APACrefYearMonthDay{2011}{}{}.
\newblock
\BBOQ{}\APACrefatitle{Clustering of gaze during dynamic scene viewing is
  predicted by motion}{Clustering of gaze during dynamic scene viewing is
  predicted by motion}.\BBCQ{}
\newblock
\APACjournalVolNumPages{Cognitive Computation}{3}{1}{5--24}.
\PrintBackRefs{\CurrentBib}

\bibitem[\protect\citeauthoryear{%
{Prof. Dr. G. Keith Still}%
}{%
{Prof. Dr. G. Keith Still}%
}{%
{\protect\APACyear{2018}}%
}]{%
crowden}%
\APACinsertmetastar{%
crowden}%
{Prof. Dr. G. Keith Still}.%
%
\unskip\
\newblock
\APACrefYearMonthDay{2018}{}{}.
\newblock
\APACrefbtitle{Crowd density - moving crowds.}{Crowd density - moving crowds.}
\newblock
 \begin{APACrefURL}
  \url{http://www.gkstill.com/Support/crowd-flow/MovingDensity.html}
  \end{APACrefURL}
\newblock
\APACrefnote{[Online; accessed 13-June-2019]}
\PrintBackRefs{\CurrentBib}

\bibitem[\protect\citeauthoryear{%
Tavakoli%
, Ahmed%
, Borji%
\BCBL{}\ \BBA{} Laaksonen%
}{%
Tavakoli%
\ \protect\BOthers{.}}{%
{\protect\APACyear{2017}}%
}]{%
tavakoli2017saliency}%
\APACinsertmetastar{%
tavakoli2017saliency}%
Tavakoli, H\BPBI R.%
, Ahmed, F.%
, Borji, A.%
\BCBL{}\ \BBA{} Laaksonen, J.%
%
\unskip\
\newblock
\APACrefYearMonthDay{2017}{}{}.
\newblock
\BBOQ{}\APACrefatitle{Saliency revisited: Analysis of mouse movements versus
  fixations}{Saliency revisited: Analysis of mouse movements versus
  fixations}.\BBCQ{}
\newblock
\BIn{} \APACrefbtitle{Proceedings of the IEEE Conference on Computer Vision and
  Pattern Recognition}{Proceedings of the ieee conference on computer vision
  and pattern recognition}\ (\BPGS\ 1774--1782).
\PrintBackRefs{\CurrentBib}

\bibitem[\protect\citeauthoryear{%
Vigier%
, Rousseau%
, Da~Silva%
\BCBL{}\ \BBA{} Le~Callet%
}{%
Vigier%
\ \protect\BOthers{.}}{%
{\protect\APACyear{2016}}%
}]{%
vigier2016new}%
\APACinsertmetastar{%
vigier2016new}%
Vigier, T.%
, Rousseau, J.%
, Da~Silva, M\BPBI P.%
\BCBL{}\ \BBA{} Le~Callet, P.%
%
\unskip\
\newblock
\APACrefYearMonthDay{2016}{}{}.
\newblock
\BBOQ{}\APACrefatitle{A new HD and UHD video eye tracking dataset}{A new hd and
  uhd video eye tracking dataset}.\BBCQ{}
\newblock
\BIn{} \APACrefbtitle{Proceedings of the 7th International Conference on
  Multimedia Systems}{Proceedings of the 7th international conference on
  multimedia systems}\ (\BPG~48).
\PrintBackRefs{\CurrentBib}

\bibitem[\protect\citeauthoryear{%
Volokitin%
, Gygli%
\BCBL{}\ \BBA{} Boix%
}{%
Volokitin%
\ \protect\BOthers{.}}{%
{\protect\APACyear{2016}}%
}]{%
volokitin2016predicting}%
\APACinsertmetastar{%
volokitin2016predicting}%
Volokitin, A.%
, Gygli, M.%
\BCBL{}\ \BBA{} Boix, X.%
%
\unskip\
\newblock
\APACrefYearMonthDay{2016}{}{}.
\newblock
\BBOQ{}\APACrefatitle{Predicting when saliency maps are accurate and eye
  fixations consistent}{Predicting when saliency maps are accurate and eye
  fixations consistent}.\BBCQ{}
\newblock
\BIn{} \APACrefbtitle{Proceedings of the IEEE Conference on Computer Vision and
  Pattern Recognition}{Proceedings of the ieee conference on computer vision
  and pattern recognition}\ (\BPGS\ 544--552).
\PrintBackRefs{\CurrentBib}

\bibitem[\protect\citeauthoryear{%
Wang%
, Shen%
, Guo%
, Cheng%
\BCBL{}\ \BBA{} Borji%
}{%
Wang%
\ \protect\BOthers{.}}{%
{\protect\APACyear{2018}}%
}]{%
wang2018revisiting}%
\APACinsertmetastar{%
wang2018revisiting}%
Wang, W.%
, Shen, J.%
, Guo, F.%
, Cheng, M\BHBI M.%
\BCBL{}\ \BBA{} Borji, A.%
%
\unskip\
\newblock
\APACrefYearMonthDay{2018}{}{}.
\newblock
\BBOQ{}\APACrefatitle{Revisiting video saliency: A large-scale benchmark and a
  new model}{Revisiting video saliency: A large-scale benchmark and a new
  model}.\BBCQ{}
\newblock
\BIn{} \APACrefbtitle{Proceedings of the IEEE Conference on Computer Vision and
  Pattern Recognition}{Proceedings of the ieee conference on computer vision
  and pattern recognition}\ (\BPGS\ 4894--4903).
\PrintBackRefs{\CurrentBib}

\bibitem[\protect\citeauthoryear{%
Yoo%
\ \protect\BOthers{.}}{%
Yoo%
\ \protect\BOthers{.}}{%
{\protect\APACyear{2016}}%
}]{%
yoo2016visual}%
\APACinsertmetastar{%
yoo2016visual}%
Yoo, Y.%
, Yun, K.%
, Yun, S.%
, Hong, J.%
, Jeong, H.%
\BCBL{}\ \BBA{} Young~Choi, J.%
%
\unskip\
\newblock
\APACrefYearMonthDay{2016}{}{}.
\newblock
\BBOQ{}\APACrefatitle{Visual path prediction in complex scenes with crowded
  moving objects}{Visual path prediction in complex scenes with crowded moving
  objects}.\BBCQ{}
\newblock
\BIn{} \APACrefbtitle{Proceedings of the IEEE Conference on Computer Vision and
  Pattern Recognition}{Proceedings of the ieee conference on computer vision
  and pattern recognition}\ (\BPGS\ 2668--2677).
\PrintBackRefs{\CurrentBib}

\end{thebibliography}
\bibliographystyle{jovcite}

\end{document}